\documentclass[]{bytedance_seed}

\usepackage[toc,page,header]{appendix}

\usepackage{minitoc}
\usepackage{wrapfig}
\usepackage{multirow}
\usepackage{makecell}
\usepackage{xcolor}
\usepackage{colortbl}
\usepackage{amsfonts}
\usepackage{float}

\definecolor{colorfirst}{rgb}{.866,.945, 0.831} % green
\definecolor{colorsecond}{rgb}{1, 0.98, 0.83} % yellow
\definecolor{colorthird}{rgb}{0.76, 0.87, 0.92} % white
\definecolor{colorcite}{rgb}{0.212, 0.490, 0.741} % white

\title{Trace Anything:\\
Representing Any Video in 4D via Trajectory Fields}

\author[1,2]{Xinhang Liu}
\author[1,3]{Yuxi Xiao}
\author[1]{Donny Y. Chen}
\author[1]{Jiashi Feng}
\author[4]{\\Yu-Wing Tai}
\author[2]{Chi-Keung Tang}
\author[1]{Bingyi Kang}

\affiliation[1]{ByteDance Seed}
\affiliation[2]{HKUST}
\affiliation[3]{Zhejiang University}
\affiliation[4]{Dartmouth College}

\abstract{
Effective spatio-temporal representation is fundamental to modeling, understanding, and predicting dynamics in videos. The atomic unit of a video, the pixel, traces a continuous 3D trajectory over time, serving as the primitive element of dynamics. Based on this principle, we propose representing any video\textsuperscript{1} as a \textit{Trajectory Field}: a dense mapping that assigns a continuous 3D trajectory function of time to each pixel in every frame. With this representation, we introduce \textit{Trace Anything}, a neural network that predicts the entire trajectory field in a single feed-forward pass. Specifically, for each pixel in each frame, our model predicts a set of control points that parameterizes a trajectory (i.e., a B-spline), yielding its 3D position at arbitrary query time instants.
We trained the Trace Anything model on large-scale 4D data, including data from our new platform, and our experiments demonstrate that:
(i) Trace Anything achieves state-of-the-art performance on our new benchmark for trajectory field estimation and performs competitively on established point-tracking benchmarks;
(ii) it offers significant efficiency gains thanks to its one-pass paradigm, without requiring iterative optimization or auxiliary estimators; and
(iii) it exhibits emergent abilities, including goal-conditioned manipulation, motion forecasting, and spatio-temporal fusion.
We will release the code, the model weights and the data platform.
}

\correspondence{Bingyi Kang}

\checkdata[Project Page]{\url{trace-anything.github.io}}
\begin{document}
\newtheorem{example}{Example}%[section]
\newtheorem{definition}{Definition}%[section]
\newtheorem{lemma}{Lemma}%[section]
\newtheorem{theorem}{Theorem}%[section]
\newtheorem{proposition}{Proposition}%[section]
\newtheorem{corollary}{Corollary}%[proposition]
\newtheorem{assumption}{Assumption}
\newtheorem{observation}{Observation}

\newcommand{\fig}[1]{Fig.~\ref{#1}}
\newcommand{\eq}[1]{Eq.~(\ref{#1})}
\newcommand{\ineq}[1]{Ineq.~(\ref{#1})}
\newcommand{\tb}[1]{Tab.~\ref{#1}}
\newcommand{\se}[1]{Section~\ref{#1}}
\newcommand{\ap}[1]{Appendix~\ref{#1}}
\newcommand{\pa}[1]{Part~\ref{#1}}
\newcommand{\lm}[1]{Lemma~\ref{#1}}
\newcommand{\prop}[1]{Proposition~\ref{#1}}
\newcommand{\alg}[1]{Algo.~\ref{#1}}
\newcommand{\theo}[1]{Theorem~\ref{#1}}
\newcommand{\defi}[1]{Definition~\ref{#1}}
\newcommand{\assum}[1]{Assumption~\ref{#1}}
\newcommand{\observe}[1]{Observation~\ref{#1}}

\newcommand*{\dif}{\mathop{}\!\mathrm{d}}
\newcommand*{\kl}{\mathrm{KL}}
\newcommand{\bbI}{\ensuremath{\mathbb{I}}} % Indicator
\newcommand{\bbE}{\ensuremath{\mathbb{E}}} % 
\newcommand{\bbS}{\ensuremath{\mathbb{S}}} % 
\newcommand{\bbR}{\ensuremath{\mathbb{R}}} % Real Numbers
\newcommand{\caA}{\ensuremath{\mathcal{A}}} % Action
\newcommand{\caS}{\ensuremath{\mathcal{S}}} % State
\newcommand{\caAt}{\ensuremath{\mathcal{\tilde{A}}}} 
\newcommand{\caSt}{\ensuremath{\mathcal{\tilde{S}}}} 
\newcommand{\caN}{\ensuremath{\mathcal{N}}} % Normal Distribution
\newcommand{\caM}{\ensuremath{\mathcal{M}}} % Model
\newcommand{\caMt}{\ensuremath{\mathcal{\tilde{M}}}}
\newcommand{\caD}{\ensuremath{\mathcal{D}}} 
\newcommand{\caG}{\ensuremath{\mathcal{G}}} 
\newcommand{\caL}{\ensuremath{\mathcal{L}}} 
\newcommand{\caT}{\ensuremath{\mathcal{T}}} 
\newcommand{\caO}{\ensuremath{\mathcal{O}}} 
\newcommand{\caTt}{\ensuremath{\mathcal{\tilde{T}}}}
\newcommand{\caB}{\ensuremath{\mathcal{B}}} 
\newcommand{\kld}{\text{D}_{\text{KL}}} 
\newcommand{\jsd}{\text{D}_{\text{JS}}}
\newcommand{\fd}{\text{D}_{\text{f}}} 
\newcommand{\iter}[2]{{#1}^{(#2)}}
\newcommand{\piE}{{\pi_E}}
\newcommand{\hr}{\hat{r}}
\newcommand{\hpi}{\hat{\pi}}

\newcommand{\hytt}[1]{\texttt{\hyphenchar\font=\defaulthyphenchar #1}}

\newcommand{\cmark}{\ding{51}}%
\newcommand{\xmark}{\ding{55}}%
\newcommand{\tc}[1]{\textcolor{red}{#1}}

\let\titleold\title
\renewcommand{\title}[1]{\titleold{#1}\newcommand{\thetitle}{#1}}
\def\maketitlesupplementary
   {
   \newpage
       \twocolumn[
        \centering
        \Large
        % \textbf{\thetitle}\\
        \textbf{Visual Whole-Body Control for Legged Loco-Manipulation}\\
        \vspace{0.5em}Appendix \\
        \vspace{1.0em}
       ] %< twocolumn
   }
\providecommand{\YJ}[1]{{\color{blue}\textbf{YJ:}{#1}}}
\maketitle

\begin{figure}[h]
    \centering
    \includegraphics[width=\linewidth]{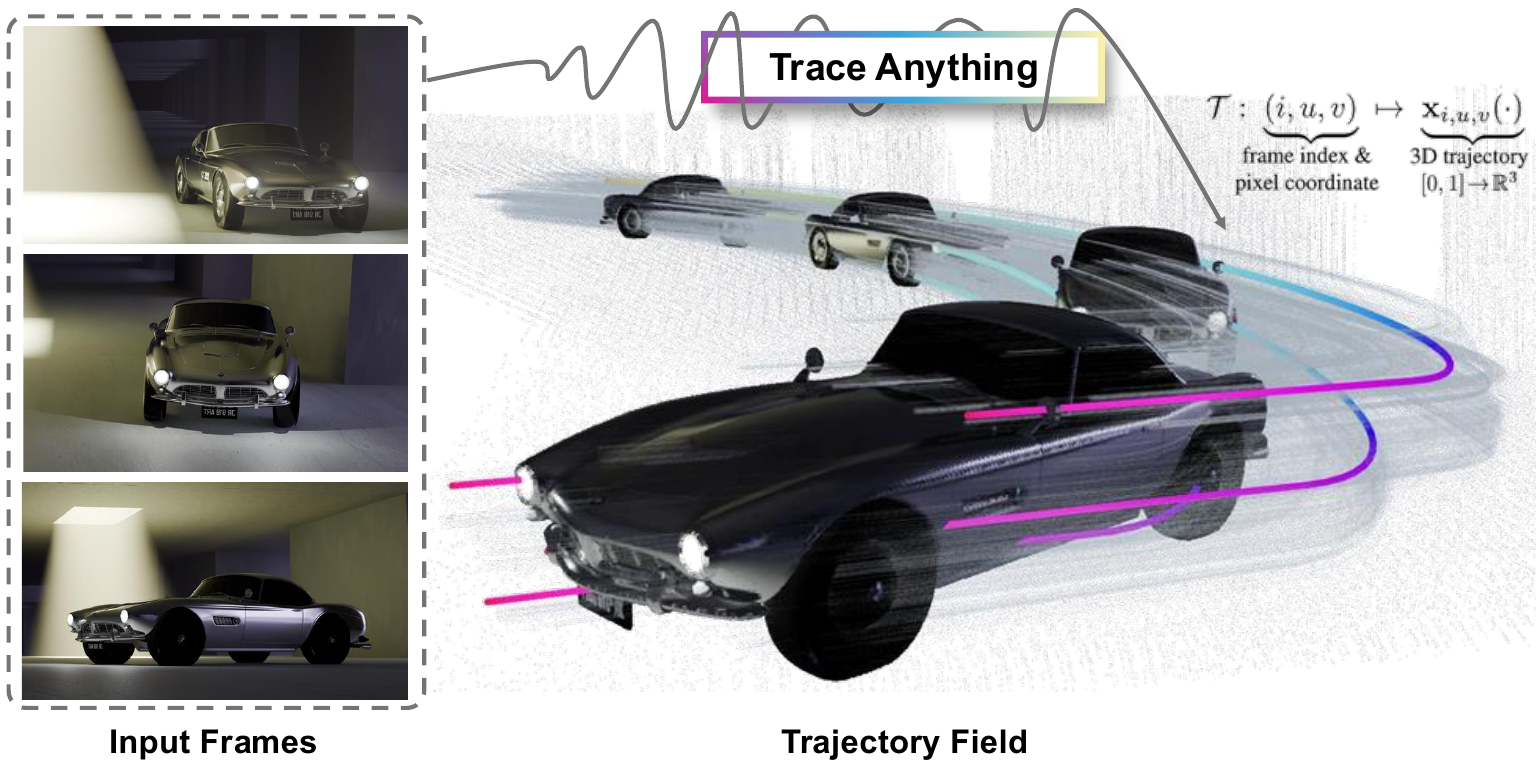}\\
\caption{Any video$^{*}$ can be represented in 4D with a \textbf{Trajectory Field}, a dense mapping assigning each pixel in each frame to a parametric 3D trajectory. We propose \textbf{Trace Anything}, a neural network that predicts the trajectory field with a single forward pass.}
 \label{fig:teaser}
\end{figure}
\footnotetext[1]{Here, ``any video'' extends beyond monocular videos to include image pairs or even unordered unstructured image collections that capture dynamic scenes.}

\section{Introduction}

\label{sec:intro}
Understanding dynamic scenes requires more than disjoint reconstruction of 3D space at each time step; it demands modeling how the scene evolves in both space and time. A central challenge 
toward spatial intelligence is to develop a 4D video representation that captures the underlying spacetime dynamics in a way that is geometrically grounded and scalable. Rather than relying on additional estimators such as depth, flow, or tracking, or on heavy per-scene optimization, we observe that the atomic elements of video, its pixels, naturally trace out \textit{3D trajectories} in the world, which acts as the atomic primitive of dynamics.

\begin{wrapfigure}{r}{0.5\linewidth} %
    \centering
    \includegraphics[width=\linewidth]{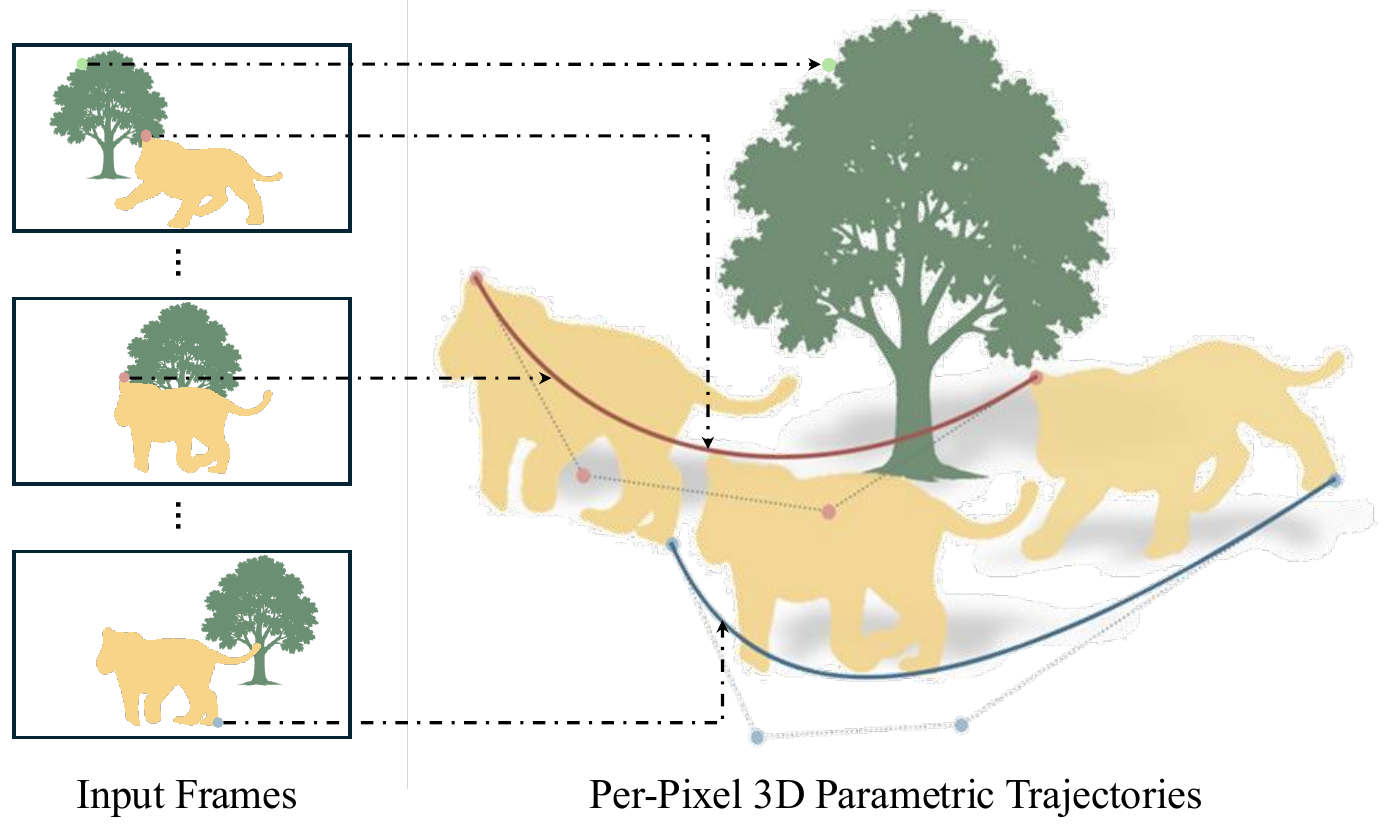}\\
    \caption{Given the input frames (left), a \textit{trajectory field} represents the video at the atomic level, mapping each pixel in each frame to a 3D trajectory, expressed as a parametric curve (right).}
    \label{fig:method_traj}
\end{wrapfigure}

Recognizing this, we propose \textbf{Trajectory Fields}, a versatile 4D representation for any video that associates each pixel in each frame with a parametric 3D trajectory, as illustrated in \Cref{fig:method_traj}. Unlike prior 4D reconstruction methods that produce disjoint per-frame point clouds~\citep{zhang2024monst3r,li2024_megasam,sucar2025dynamic} and rely on estimated optical flow or 2D tracks to build cross-frame correspondences, Trajectory Fields offer a more direct and compact way to model scene dynamics.

Building on this representation, we propose \textbf{Trace Anything}, a feed-forward neural network that estimates trajectory fields directly from video frames. 
As shown in \Cref{fig:teaser}, with a single forward pass over all input frames, it predicts a stack of control point maps for each frame, defining spline-based parametric trajectories for every pixel. 
This design brings three key advantages: (i) its one-pass scheme eliminates intermediate estimators and iterative global alignment, (ii) it predicts all trajectories (per pixel per frame) jointly in a shared world coordinate system, and (iii) it generalizes across diverse inputs, including monocular videos, image pairs, and unordered photo sets.

To support training and evaluation at scale, we develop a Blender-based platform featuring diverse environments, moving characters, and camera trajectories. It produces photo-realistic renderings with dense annotations, including 2D/3D trajectories, depth, semantics, flow, and camera poses. From this platform, we release (i) the \textit{Trace Anything dataset}—10,000+ videos (120 frames each) for training trajectory field estimation models, and (ii) the \textit{Trace Anything benchmark}—200 curated videos for evaluating models’ ability to capture motion jointly across all frames.

Trained with our new dataset, Trace Anything achieves state-of-the-art results on our trajectory field benchmark and performs competitively on established point tracking benchmarks, while offering significant efficiency gains. Moreover, our paradigm also reveals new capabilities for spatial reasoning, including motion forecasting, spatio-temporal fusion, and goal-conditioned manipulation.

In summary, our contributions are:
\begin{itemize}
    \item We propose \textbf{Trajectory Fields} as an atomic-level and versatile 4D video representation, grounded in a principled formulation.
    \item We present \textbf{Trace Anything}, a feed-forward network that predicts trajectory fields without requiring extra estimators or per-scene optimization.
    \item We develop a synthetic data platform for large-scale training and benchmarking of trajectory field estimation.
    \item Extensive experiments on existing and new benchmarks demonstrate competitive accuracy, faster inference, and new capabilities.
\end{itemize}

\section{Related Work}
\label{sec:related_work}

\noindent\textbf{(Dynamic) 3D scene reconstruction.}  
Reconstructing 3D structure from multi-view images is a long-standing problem in computer vision. 
Classical Structure-from-Motion (SfM) pipelines~\citep{hartley2003multiple, agarwal2011building, schonberger2016structure} proceed in sequential stages: feature extraction, image matching, triangulation, relative pose estimation, and global bundle adjustment. 
Deep learning has improved individual components~\citep{detone2018superpoint, sarlin2020superglue} yet stage-wise pipelines remain prone to error accumulation.  
DUSt3R~\citep{wang2024dust3r} addressed this by directly predicting 3D pointmaps from image pairs. 
Fast3R~\citep{fast3r}, VGGT~\citep{wang2025vggt}, $\pi^3$~\citep{wang2025pi3} and MapAnything~\citep{keetha2025mapanything} further relaxed the pairwise assumption with all-to-all attention, enabling joint reasoning over all frames and avoiding $O(N^2)$ pairwise inference.  
However, both traditional and learning-based methods generally assume static scenes and sufficient camera baselines, leading to degraded performance in dynamic settings.  
To handle monocular videos with dynamics, MegaSAM~\citep{li2024_megasam} integrates optimization-based SLAM, while Monst3R~\citep{zhang2024monst3r}, POMATO~\citep{zhang2025pomato}, Easi3R~\cite{chen2025easi3r}, St4RTrack~\citep{feng2025st4rtrack}, and Dynamic Point Maps~\citep{sucar2025dynamic} extend DUSt3R-style networks to dynamic scenes.  
These methods typically generate disjoint per-frame point clouds, relying on optical flow or 2D tracks for cross-frame correspondences, and their pairwise inference often requires costly per-scene optimization for global alignment.  
In contrast, Trace Anything directly estimates trajectory fields that produce dynamic point clouds with cross-frame correspondences, sharing the feed-forward spirit of~\citet{fast3r} and \cite{wang2025vggt} and performing one-pass inference over all frames.

\noindent\textbf{Point tracking.} 
Particle Video~\citep{sand2008particle} first introduced long-range particle trajectories in videos. 
Early deep learning methods~\citep{harley2022particle, doersch2022tap, doersch2023tapir} approached this with global matching and local refinement. 
CoTracker~\citep{karaev2024cotracker} leveraged a transformer-based architecture to enable tracking through occlusions, followed by works~\citep{li2024taptr, cho2024local} that improved efficiency with 4D correlation volumes. 
CoTracker3~\citep{karaev2024cotracker3} further leveraged unlabeled data to boost performance.  
3D point tracking remains comparatively new. 
OmniMotion~\citep{wang2023tracking} addressed the task via test-time optimization, while SpatialTracker~\citep{xiao2024spatialtracker} introduced the first feed-forward 3D tracker by combining 2D tracking with monocular depth priors. 
DELTA~\citep{ngo2024delta} achieved dense 3D tracking using a transformer with upsampling for high-resolution outputs. 
Concurrently, SpatialTrackerV2~\citep{xiao2025spatialtrackerv2} scaled training across real and synthetic data, and St4RTrack~\citep{feng2025st4rtrack} and POMATO~\cite{zhang2025pomato} extended 3D reconstruction models for tracking via joint optimization.  
Unlike prior approaches, our method bypasses monocular depth estimation and 2D trackers and directly predicts dense 3D trajectories in a feed-forward manner.

\noindent\textbf{4D representations for NVS.}  
A large class of 4D representations has been developed for novel view synthesis (NVS) in dynamic scenes, aiming to deliver immersive effects such as “bullet time.” Since Neural Radiance Fields (NeRF)~\citep{mildenhall2020nerf} introduced implicit volumetric representations, many extensions have incorporated temporal dynamics. One class of methods~\citep{gao2021dynamic, li2021neural, li2022neural, xian2021space} directly conditions the radiance field on time, treating density and color as continuous functions of space and time. Another class~\citep{pumarola2021d, zhang2021editable, park2021nerfies, park2021hypernerf, tretschk2021non} maps observations to a canonical space and models dynamics via deformation fields. Grid-based approaches~\citep{cao2023hexplane, kplanes, wang2022fourier, attal2023hyperreel, liu2024gear} discretize the 4D volume into compact planar factors for efficiency. Also in this line of work, \citet{wang2021neural} proposed `neural trajectory fields', with a different formulation and purpose than `trajectory fields' in this study. More recently, 3D Gaussian Splatting (3DGS)~\citep{3dgs} has been extended to dynamics~\citep{wu20234d, yang2023real, luiten2023dynamic, yang2023deformable, li2024spacetime}, improving rendering quality and speed. These efforts focus on photorealistic appearance and typically assume precomputed camera poses or point clouds. Our work is orthogonal: we propose a geometry-centric paradigm that directly infers trajectory fields from raw videos, emphasizing accurate 3D motion modeling. Integrating NVS with our paradigm, e.g., using trajectory fields to initialize dynamic 3DGS models, is a promising future direction.

\begin{figure*}[t]
    \centering
    \includegraphics[width=\linewidth]{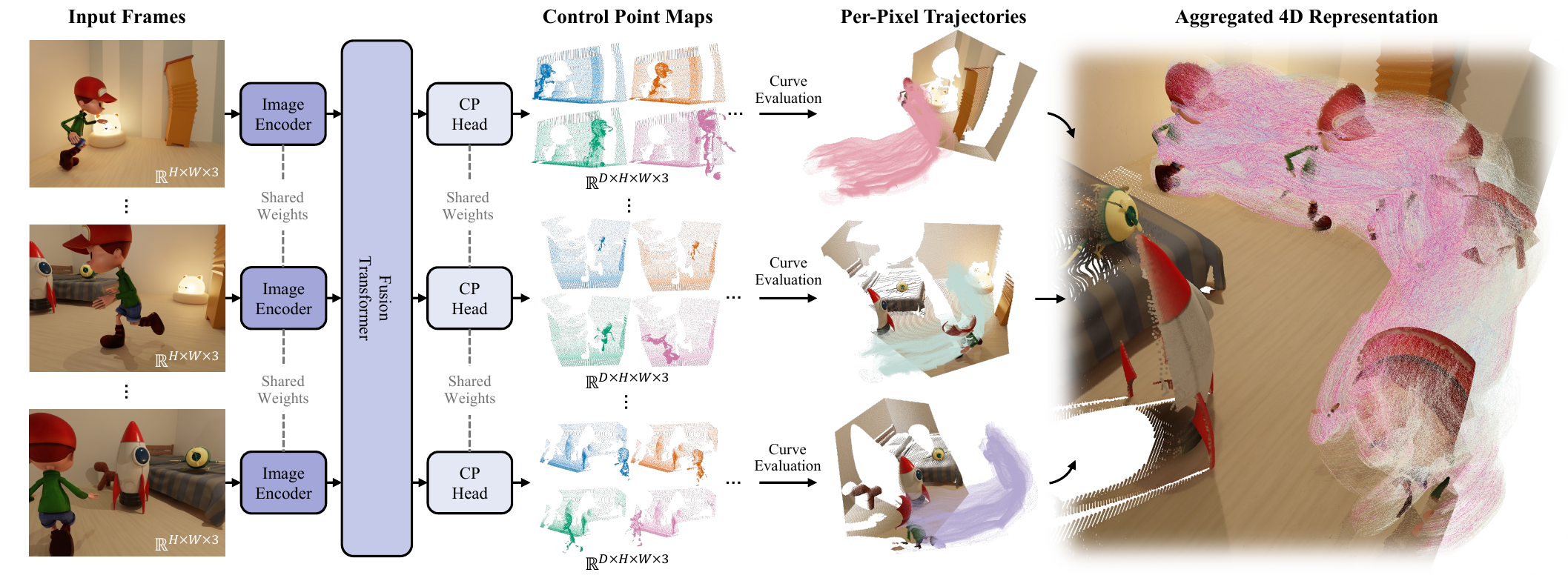}\\
    \caption{\textbf{Trace Anything pipeline.} Input frames are processed by a geometric backbone consisting of an image encoder and a fusion transformer. The control point head outputs dense control point maps \(\mathbf{P}_i \in \mathbb{R}^{D \times H \times W \times 3}\), where \(\mathbf{P}^{(k)}_{i,u,v}\) is the \( k \)-th control point for pixel \((u,v)\) in frame \( I_i \). These define continuous 3D trajectories \(\mathbf{x}_{i,u,v}(t)\) via cubic B-splines, yielding a 4D reconstruction.}
    \label{fig:method_arch}
\end{figure*}

\section{Method}
\label{sec:methods}
The atomic elements of video, its pixels, naturally trace out 3D trajectories in the world, forming the primitive units of dynamics. 
Recognizing this, we aim to model dynamic scenes through trajectory fields, a 4D representation that encodes each pixel in each frame as a continuous 3D trajectory over time. In the following, we first formalize trajectory fields in \Cref{sec:problem}, then present Trace Anything in \Cref{sec:network}, a feed-forward network designed to estimate them, and finally describe the overall training scheme in \Cref{sec:training}.
In this section, we define a \textit{field} as a mapping from a domain \(M\) to a codomain \(V\), \(\mathcal{F}: M \to V\), where \(M\) may be a discrete or continuous space, and \(V\) may represent scalars, vectors, or functions. We provide preliminaries on fields in \Cref{supp:field} and on parametric curves in \Cref{supp:parametric_curves}.

\subsection{Problem Formulation}
\label{sec:problem}
We formalize \textit{trajectory fields}, a 4D representation of dynamic 3D scenes in a video. Let \(\{I_i\}_{i=1}^N\) be a collection of \( N \) RGB frames, where each \( I_i \in \mathbb{R}^{3 \times H \times W} \) captures the scene at different time steps or viewpoints. A trajectory field is defined as
\begin{equation}
\mathcal{T}: [N] \times [H] \times [W] \;\to\; C([0,1], \mathbb{R}^3), \quad
(i, u, v) \;\mapsto\; \mathbf{x}_{i,u,v}(\cdot)
\end{equation}
where \([N]\), \([H]\), and \([W]\) denote the discrete sets of frame indices and pixel coordinates, respectively, and \(\mathbf{x}_{i,u,v} : [0,1] \to \mathbb{R}^3\) is a continuous 3D trajectory for pixel \((u,v)\) in frame \( I_i \). 
The domain is \( M = [N] \times [H] \times [W] \), and the codomain is \( V = C([0,1], \mathbb{R}^3) \), the space of continuous functions from \([0,1]\) to \(\mathbb{R}^3\). 
Each trajectory \(\mathbf{x}_{i,u,v}(t)\) is parameterized as a spline-based curve with \( D \) control points, defined as
\begin{equation}
    \mathbf{P}_i \in \mathbb{R}^{D \times H \times W \times 3},
\end{equation}
where \(\mathbf{P}^{(k)}_{i,u,v} \in \mathbb{R}^3\) is the \( k \)-th control point for pixel \((u,v)\) in frame \( I_i \), with \( k \in \{0, 1, \dots, D-1\} \). Given basis functions \(\{\phi_k(t)\}_{k=0}^{D-1}\), the trajectory is
\begin{equation}
    \mathbf{x}_{i,u,v}(t) = \sum_{k=0}^{D-1} \mathbf{P}^{(k)}_{i,u,v} \phi_k(t).
    \label{eq:trajectory}
\end{equation}
The form of the basis functions \(\{\phi_k(t)\}_{k=0}^{D-1}\) depends on the type of parametric curve. In our implementation, we use cubic B-splines with clamped knots as detailed in \Cref{supp:parametric_curves}.

\Cref{fig:method_traj} illustrates this formulation of trajectory fields. For any pixel from any frame, its 3D coordinate at any time $ t \in [0,1] $ can be obtained with \Cref{eq:trajectory}. This fundamentally differs from existing 4D reconstruction methods that predict per-frame disjoint point clouds and establish cross-frame correspondences via estimated optical flow or 2D tracks.
Ideally, two conditions should hold:
(C1) pixels in static regions collapse to degenerate trajectories;
(C2) corresponding pixels from different frames map to the same 3D trajectory.

\subsection{Network Architecture}
\label{sec:network}
Building on the formulation in \Cref{sec:problem}, we propose a feed-forward network, \textit{Trace Anything} (\Cref{fig:method_arch}), which predicts trajectory fields directly from video or unstructured image sets.  
For each frame, it outputs control point maps defining parametric curves over time, enabling trajectory field estimation in a single pass.  
This design eliminates reliance on depth or optical flow and avoids per-scene iterative optimization, providing a compact, efficient approach to modeling 4D scenes.

\noindent\textbf{Geometric backbone.}  
We build Trace Anything upon a feed-forward geometric backbone, similar in spirit to recent models~\citep{wang2025vggt, fast3r}. 
Each frame is first tokenized by an image encoder, followed by a fusion transformer that integrates spatio-temporal context across views through interleaved frame-wise and global attention layers.
For sequential video input, we additionally inject temporal index embeddings, while the architecture remains compatible with unordered or unstructured image collections.

\noindent\textbf{Control Point Head.}  
Built on the backbone features, the \textit{control point head} outputs dense control point maps $\mathbf{P}_i \in \mathbb{R}^{D \times H \times W \times 3}$ for each input frame $I_i$.  
Each pixel $(u,v)$ has $D$ control points $\{\mathbf{P}^{(k)}_{i,u,v}\}_{k=0}^{D-1}$, compactly parameterizing its 3D trajectory.  
Predictions are in a shared world coordinate system, with an optional \textit{local CP head} estimating control points in each frame’s local camera system.  
The head also predicts \textit{per-control-point confidence scores} $\Sigma^{(k)}_{i,u,v}$ for confidence-weighted training and filtering uncertain estimates at inference.

%On top of the backbone features, the control point head outputs a stack of dense control point maps $\mathbf{P}_i \in \mathbb{R}^{D \times H \times W \times 3}$ for each input frame $I_i$.  
%Each pixel $(u,v)$ is assigned $D$ control points $\{\mathbf{P}^{(k)}_{i,u,v}\}_{k=0}^{D-1}$, which compactly parameterize its 3D trajectory.  
%The predictions are primarily expressed in a shared world coordinate system across all images, with an additional head estimating control points in each frame’s local camera coordinate system.
%In addition, the head predicts a per-control-point confidence score $\Sigma^{(k)}_{i,u,v}$, which is used both for confidence-weighted training losses and for filtering uncertain estimates during inference.

\noindent\textbf{Curve evaluation.}  
Given the predicted control points and basis functions $\{\phi_k(t)\}_{k=0}^{D-1}$, continuous trajectories $\mathbf{x}_{i,u,v}(t)$ are obtained via \Cref{eq:trajectory}.  
At evaluation time, the trajectory can be queried at any $t \in [0,1]$. In particular,
\begin{equation}
\mathbf{x}_{i,u,v}(0) = \sum_{k=0}^{D-1} \mathbf{P}^{(k)}_{i,u,v} \cdot \phi_k(0) \;\overset{*}{=}\; \mathbf{P}^{(0)}_{i,u,v}, \quad
\mathbf{x}_{i,u,v}(1) = \sum_{k=0}^{D-1} \mathbf{P}^{(k)}_{i,u,v} \cdot \phi_k(1) \;\overset{*}{=}\; \mathbf{P}^{(D-1)}_{i,u,v},
\end{equation}
where $(*)$ holds for cubic B-splines with clamped knots or for Bézier bases.

To obtain the 3D coordinates of a pixel from frame $i$ evaluated at the acquisition time of another frame $j$, we substitute the corresponding temporal parameter $t_j$ into its trajectory:
\begin{equation}
    \mathbf{X}_{i \rightarrow j}(u,v) = \mathbf{x}_{i,u,v}(t_j).
    \label{eqn:i2j}
\end{equation}
In most cases, $t_j$ is obtained from metadata or frame order. Otherwise, an auxiliary \textit{timestamp head} predicts normalized timestamps $\hat{t}_j \in [0,1]$. 
As a special case, evaluating each trajectory at frame $i$'s own acquisition time $t_i$ recovers the 3D point map for frame $I_i$:
\begin{equation}
\label{eqn:self_pointmap}
    \mathbf{X}_i(u,v) = \mathbf{x}_{i,u,v}(t_i).
\end{equation}

Trace Anything outputs the trajectory field with a single network inference for all frames, avoiding pairwise inference and subsequent global alignment, while being self-contained and independent of external estimators for monocular depth, optical flow, or 2D tracks.

\subsection{Training Scheme}
\label{sec:training}
To train Trace Anything, we directly supervise the accuracy of predicted trajectories.
Intuitively, a trajectory predicted from frame $i$ should, when evaluated at the timestamp of another frame $j$, land exactly at its ground-truth 3D location at frame $j$'s acquisition time. 

\noindent\textbf{Trajectory loss.}  
For a pixel $(u,v)$ in frame $i$, the predicted 3D position evaluated at $t_j$ is $\mathbf{X}_{i \rightarrow j}(u,v)$ (\Cref{eqn:i2j}), while the corresponding ground truth is $\mathbf{X}^{\text{gt}}_{i \rightarrow j}(u,v)$.  
We define the loss as
\begin{equation}
    \ell_{i \rightarrow j}(u,v) = 
    \big\| \mathbf{X}_{i \rightarrow j}(u,v) - \mathbf{X}^{\text{gt}}_{i \rightarrow j}(u,v) \big\|_2^2.
    \label{eq:traj-loss}
\end{equation}
\noindent\textbf{Confidence adjustment.}  
To account for the varying reliability of predicted trajectories across pixels and control points, we incorporate confidence adjustment.  
For each control point, the network predicts a scalar confidence $\hat\Sigma^{(k)}_{i,u,v} > 0$ alongside its 3D coordinates.  
The confidence associated with $\mathbf{X}_{i \rightarrow j}(u,v)$ is then aggregated using the same basis functions as in \Cref{eq:trajectory}:
\begin{equation}
    \hat\Sigma_{i \rightarrow j}(u,v) =
    \sum_{k=0}^{D-1} \hat\Sigma^{(k)}_{i,u,v} \cdot \phi_k(t_j).
    \label{eq:sigma-agg}
\end{equation}
The final confidence-adjusted loss then becomes
\begin{equation}
\begin{aligned}
    \mathcal{L}_{\text{traj-conf}}
    = \frac{1}{|\Omega|} \sum_{(i,j)} \sum_{(u,v)\in\Omega}
    \Big[ \, \hat\Sigma_{i \rightarrow j}(u,v)\,\ell_{i \rightarrow j}(u,v) + \alpha \log \hat\Sigma_{i \rightarrow j}(u,v) \Big],
\end{aligned}
\label{eq:traj-conf-loss}
\end{equation}
where $\Omega$ denotes valid pixels with ground-truth supervision.  
This adjustment downweights uncertain predictions while discouraging overconfident ones.

\noindent\textbf{Timestamp supervision.}  
When ground-truth timestamps are available, we directly supervise  \textit{Timestamp Head} with an $L_1$ regression loss:
\begin{equation}
\mathcal{L}_{\text{time}}
= \frac{1}{N} \sum_{i=1}^N \big| \hat{t}_i - t_i \big|.
\end{equation}

\noindent\textbf{Static regularization.}  
To encourage condition (C1), pixels in static regions should map to overlapped 3D control points. We enforce this by minimizing the variance of their control points:
\begin{equation}
\mathcal{L}_{\text{static}}
= \frac{1}{|\Omega_{\text{static}}|}
\sum_{(i,u,v)\in\Omega_{\text{static}}}
\text{Var}\!\left(\{\mathbf{P}^{(k)}_{i,u,v}\}_{k=0}^{D-1}\right).
\end{equation}

\noindent\textbf{Rigidity regularization.}  
For pixels segmented as belonging to the same rigid region, their trajectories should preserve internal distances across control points. 
Equivalently, the pairwise distance between any two pixels $p,q$ within a rigid segment should remain constant across control points.  
We enforce this by minimizing the variance of their distances:  
\begin{equation}
\mathcal{L}_{\text{rigid}}
= \frac{1}{|\Omega_{\text{rigid}}|}
\sum_{(p,q)\in\Omega_{\text{rigid}}}
\text{Var}\!\left(\big\{\|\mathbf{P}^{(k)}_{p} - \mathbf{P}^{(k)}_{q}\|_2\big\}_{k=0}^{D-1}\right).
\end{equation}

\noindent\textbf{Correspondence regularization.}  
To encourage condition (C2), pixels with known cross-frame correspondences should share identical control points.  
Let $\Omega_{\text{corr}}$ be the set of matched pixels $((i,u,v),(j,u',v'))$.  
We penalize discrepancies between their control-point sequences:
\begin{equation}
\mathcal{L}_{\text{corr}}
= \frac{1}{|\Omega_{\text{corr}}|}
\sum_{\substack{((i,u,v),\\(j,u',v')) \in \Omega_{\text{corr}}}}
\frac{1}{D}\sum_{k=0}^{D-1}
\big\|
\mathbf{P}^{(k)}_{i,u,v} - \mathbf{P}^{(k)}_{j,u',v'}
\big\|_2^2.
\end{equation}

\noindent\textbf{Final objective.}  
The overall loss combines the core trajectory supervision with the above regularization terms:
\begin{equation}
\begin{split}
\mathcal{L}
= \mathcal{L}_{\text{traj-conf}}
+ \lambda_{\text{time}} \mathcal{L}_{\text{time}}
+ \lambda_{\text{static}} \mathcal{L}_{\text{static}}
+ \lambda_{\text{rigid}} \mathcal{L}_{\text{rigid}}
+ \lambda_{\text{corr}} \mathcal{L}_{\text{corr}}.
\end{split}
\end{equation}

\begin{figure}[t]
\centering
\includegraphics[width=\linewidth]{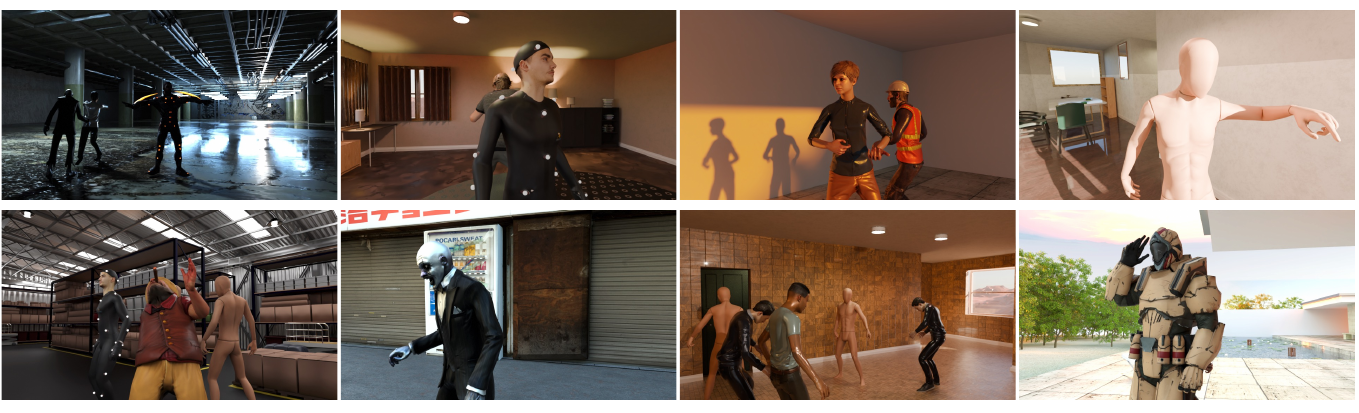}
\caption{Sample renderings from our data platform.}
\label{fig:benchmark}
\end{figure}
\section{Trace Anything Data Platform}
\label{sec:benchmark}

Data-driven modeling of dynamic scenes is limited by the lack of large-scale datasets with dense, reliable annotations. Existing synthetic datasets and generators~\citep{dosovitskiy2015flownet,mayer2016large,harley2022particle,greff2021kubric,zheng2023pointodyssey} are typically small and biased toward rigid motion, with sparse or short-term annotations, which are insufficient for realistic scene understanding and diverse dynamics. To address this, we develop a scalable 4D Scene Data Platform in Blender that synthesizes photo-realistic dynamic scenes with dense ground-truth annotations.

\noindent\textbf{Trace Anything dataset.} Using our platform, we build a dataset whose primary purpose is to train the Trace Anything model on trajectory field estimation. The current release contains about 10K unique scenes, each with 120 annotated frames. The collection spans a wide range of settings and motions, with examples shown in \Cref{fig:benchmark}. The dataset exhibits diversity across multiple aspects:
\textit{(i) Environment} — diverse indoor and outdoor backgrounds from public asset libraries and procedural generation~\citep{infinigen2023infinite,infinigen2024indoors};
\textit{(ii) Dynamics} — articulated human characters and movable objects with both rigid and non-rigid motion;
\textit{(iii) Camera motion} — smooth trajectories sampled around active regions to mimic natural filming.
Rendered RGB videos are paired with per-pixel 2D/3D trajectories, depth maps, camera poses, semantic masks, which facilitate the training scheme introduced in \Cref{sec:training}. Since our platform is fully programmable, it can be easily extended with new assets, domains, or annotation modalities to support future research.

\noindent\textbf{Trace Anything benchmark.}
To evaluate the task of trajectory field estimation, we construct a benchmark consisting of 200 videos, each with 120 frames. A key difference from established point tracking datasets~\citep{koppula2024tapvid} lies in the evaluation protocol: point tracking benchmarks evaluate estimated trajectories only for pixels sampled from the first frame \textit{(first-to-all)}, whereas our benchmark evaluates trajectories for pixels sampled from \textit{all} frames \textit{(all-to-all)}. This requires models not only to follow motion from a single starting frame, but also to jointly capture dynamics across the entire sequence. In addition, our benchmark provides denser trajectory annotations, covering more pixels per framel, and evaluates in world coordinates, requiring models to reason about both global geometry and motion.
\section{Experiments}
\label{sec:experiments}
In this section, we evaluate our method across a series of challenging settings, 
demonstrating its competitive accuracy, faster inference, and novel capabilities. 
Please refer to the appendix for additional experimental results, and to our project page for videos and interactive demos.

\subsection{Experimental Details}
We generate training data using Kubric~\citep{greff2021kubric} and our proposed 4D scene data engine. Specifically, we render 20K videos with 24 frames each using Kubric, with half containing continuous camera motion and the other half discrete camera motion, and over 10K videos with 120 frames each from our engine. While Kubric equips models with preliminary ability to capture rigid object motion, it is largely limited to rigid dynamics and textured backgrounds. Our engine complements this with diverse non-rigid object motions and more complex, varied environments.

Our released model uses an image encoder and fusion transformer initialized with Fast3R~\citep{fast3r} pretrained weights, while the CP heads are randomly initialized. For the choice of parametric curves, our released model adopts B-splines, as detailed in \Cref{supp:parametric_curves}.
All models are trained on images at a resolution of 512 pixels on the longest side, using AdamW~\citep{loshchilov2017fixing} with a learning rate of 0.0001 and a cosine annealing schedule. 

In the first stage, we train on 20K Kubric videos; in the second stage, we use a mixture of 20K Kubric videos and 10K from our engine. We adopt a batch size of 1, with each batch sampling up to 30 frames. Training takes 7.22 days on 32 NVIDIA A100 80GB GPUs. To enable efficient large-scale training, we leverage FlashAttention~\citep{dao2022flashattention,dao2023flashattention} for improved time and memory efficiency, and apply DeepSpeed ZeRO Stage 2~\citep{rajbhandari2020zero}, which partitions optimizer states, moment estimates, and gradients across machines.

\subsection{Trajectory Field Estimation}
We present qualitative results of trajectory field estimation on videos and image pairs. Qualitative comparisons are provided in \Cref{sec:davis_comparison}.

\begin{figure*}[t]
    \centering
    \includegraphics[width=\linewidth]{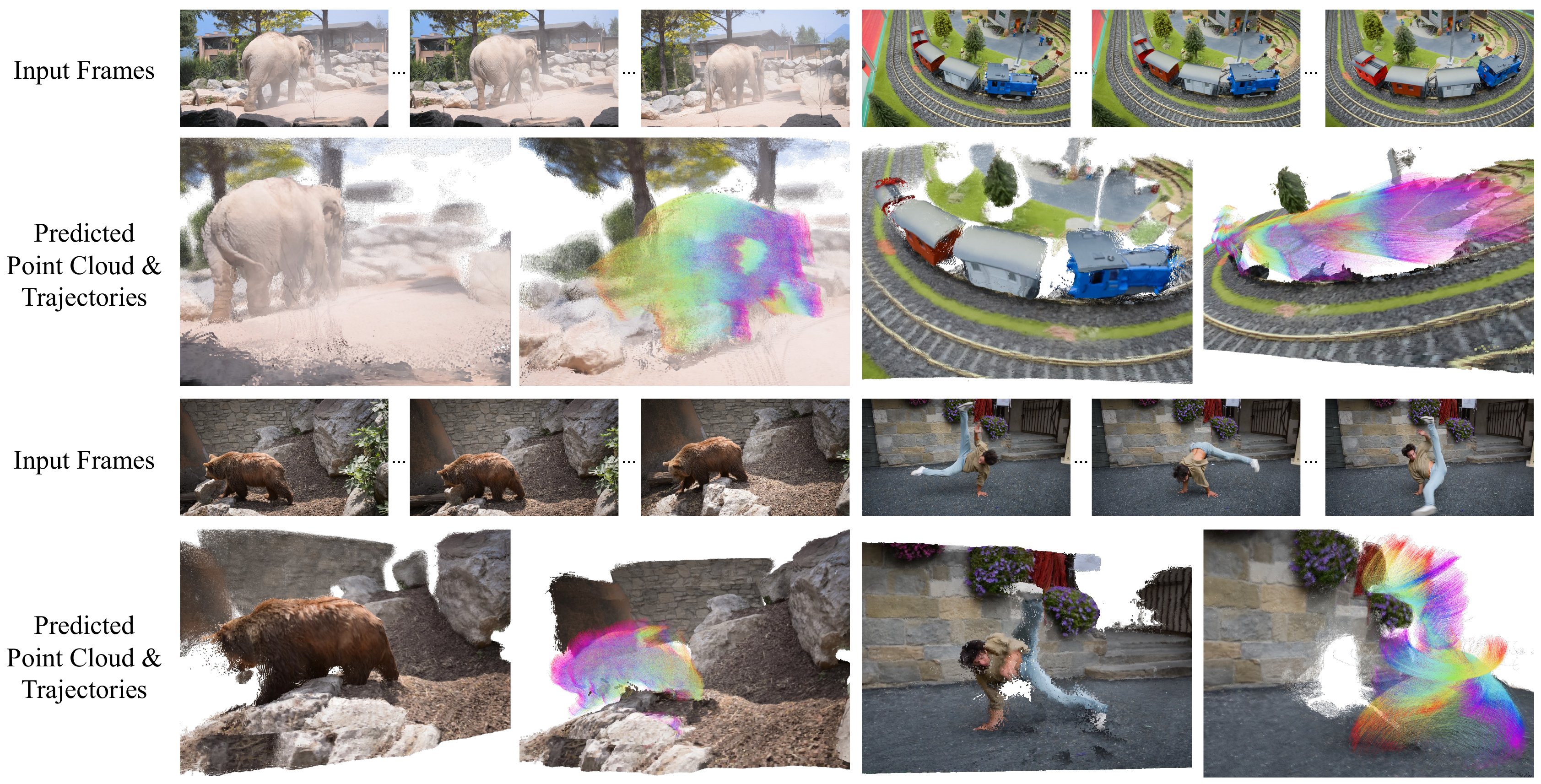}\\
    \caption{\textbf{Video-based trajectory field estimation on DAVIS~\citep{perazzi2016benchmark}.} 
    Trace Anything predicts trajectory fields that can yield dynamic point cloud sequences and dense 3D trajectories, while remaining robust to complex non-rigid motion and occlusions.}
    \label{fig:davis}
\end{figure*}

\noindent\textbf{Video to trajectory field.}  
For computational efficiency, we uniformly subsample long sequences to fewer than 60 frames. Figure~\ref{fig:davis} shows qualitative results on DAVIS~\citep{perazzi2016benchmark}, covering diverse dynamic scenes. Our predictions faithfully reconstruct both dynamic and static components of the scene, yielding dense, pixel-level 3D trajectories. These trajectories capture motions ranging from near-rigid transformations, such as a toy train moving along a track, to highly non-rigid deformations, such as humans or animals in motion, while also handling severe occlusions and preserving global scene structure.

\begin{figure}[t]
    \centering
    \includegraphics[width=0.9\linewidth]{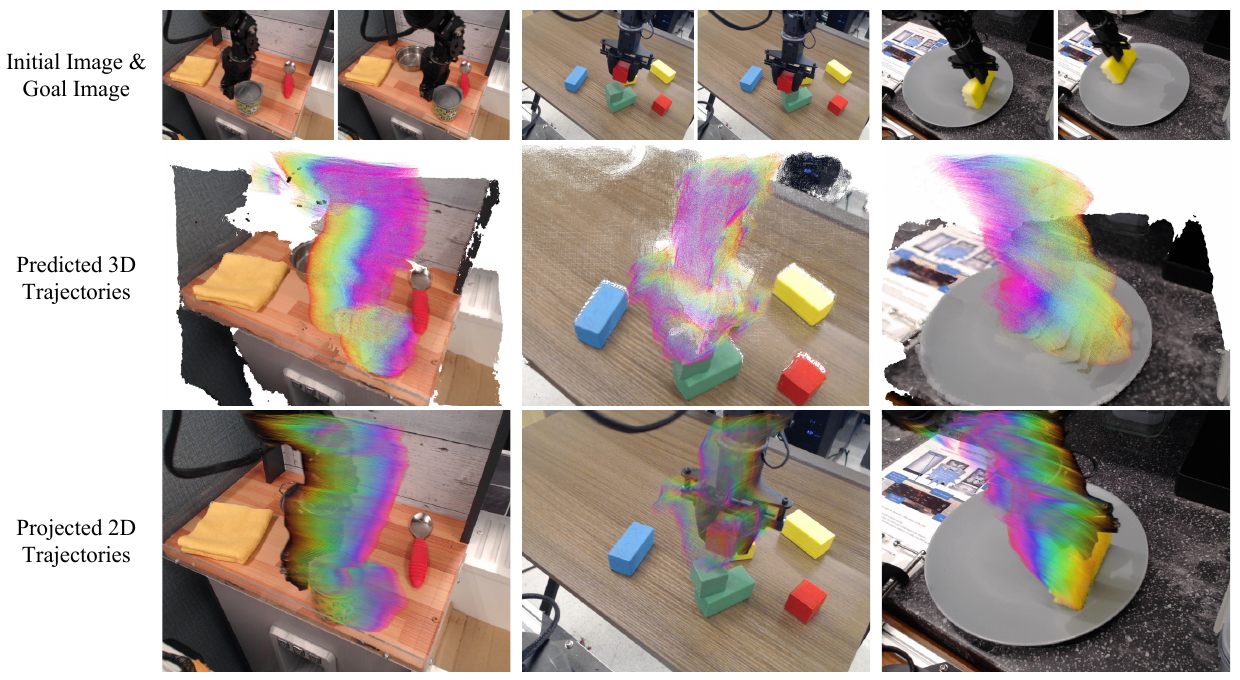}\\
    \caption{\textbf{Image-pair-based trajectory field estimation (goal-conditioned manipulation) on Bridge~\citep{walke2023bridgedata}.} 
    Given an initial and a goal image, Trace Anything predicts a trajectory field that interpolates the 3D motion of both the robot arm and manipulated objects. We further show the projected 2D trajectories (see \Cref{supp:2dtraj_etc} and \Cref{fig:2dtraj} for details).}
    \label{fig:goal}
\end{figure}

\begin{figure}[H]
    \centering
    \includegraphics[width=0.9\linewidth]{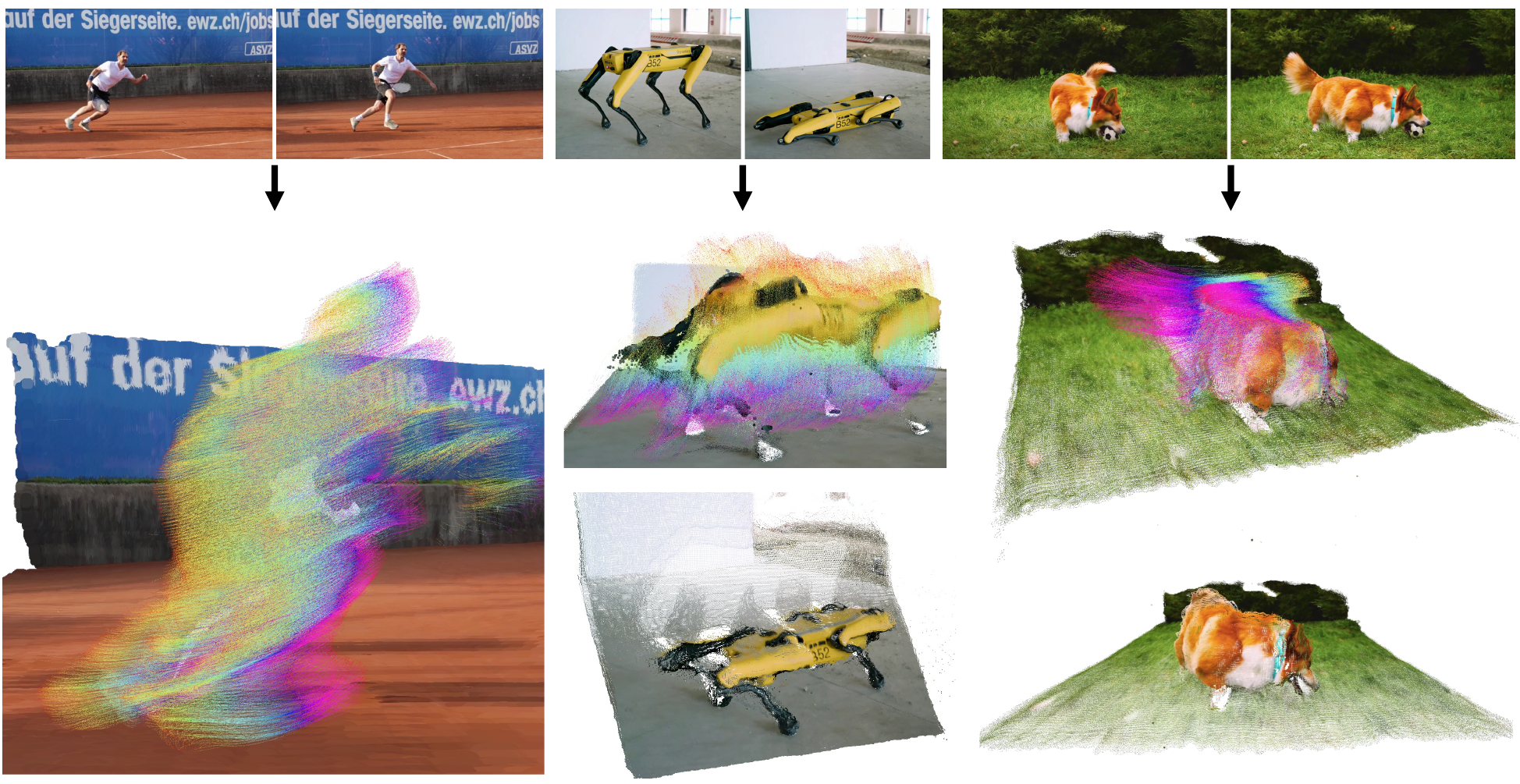}
    \caption{Qualitative results with image pairs as input.}
    \label{fig:additional_pair}
\end{figure}

\noindent\textbf{Image pair to trajectory field.}  
Our approach can also infer trajectory fields directly from image pairs, effectively reconstructing the implied spatio-temporal dynamics and interpolating intermediate motion. For this experiment, we use BridgeData V2~\citep{walke2023bridgedata}, a large and diverse dataset of robotic manipulation behaviors. Image pairs are sampled from video sequences with a temporal gap of 10–20 frames. As illustrated in \Cref{fig:goal} and \Cref{fig:additional_pair}, given an initial image of a scene and a goal image specifying the desired outcome, our model predicts a trajectory field that captures plausible 3D trajectories of both objects and agents involved. These trajectories can also be re-projected with estimated camera poses to yield 2D trajectories (see \Cref{supp:2dtraj_etc} and \Cref{fig:2dtraj} for details). In the context of robot learning, this naturally aligns with \textit{goal-conditioned manipulation}, where predicted trajectories can be interpreted as feasible robot end-effector motions~\citep{bharadhwaj2024track2act}.

\begin{figure}[t]
\vspace{-1cm}
    \centering
    \includegraphics[width=0.8\linewidth]{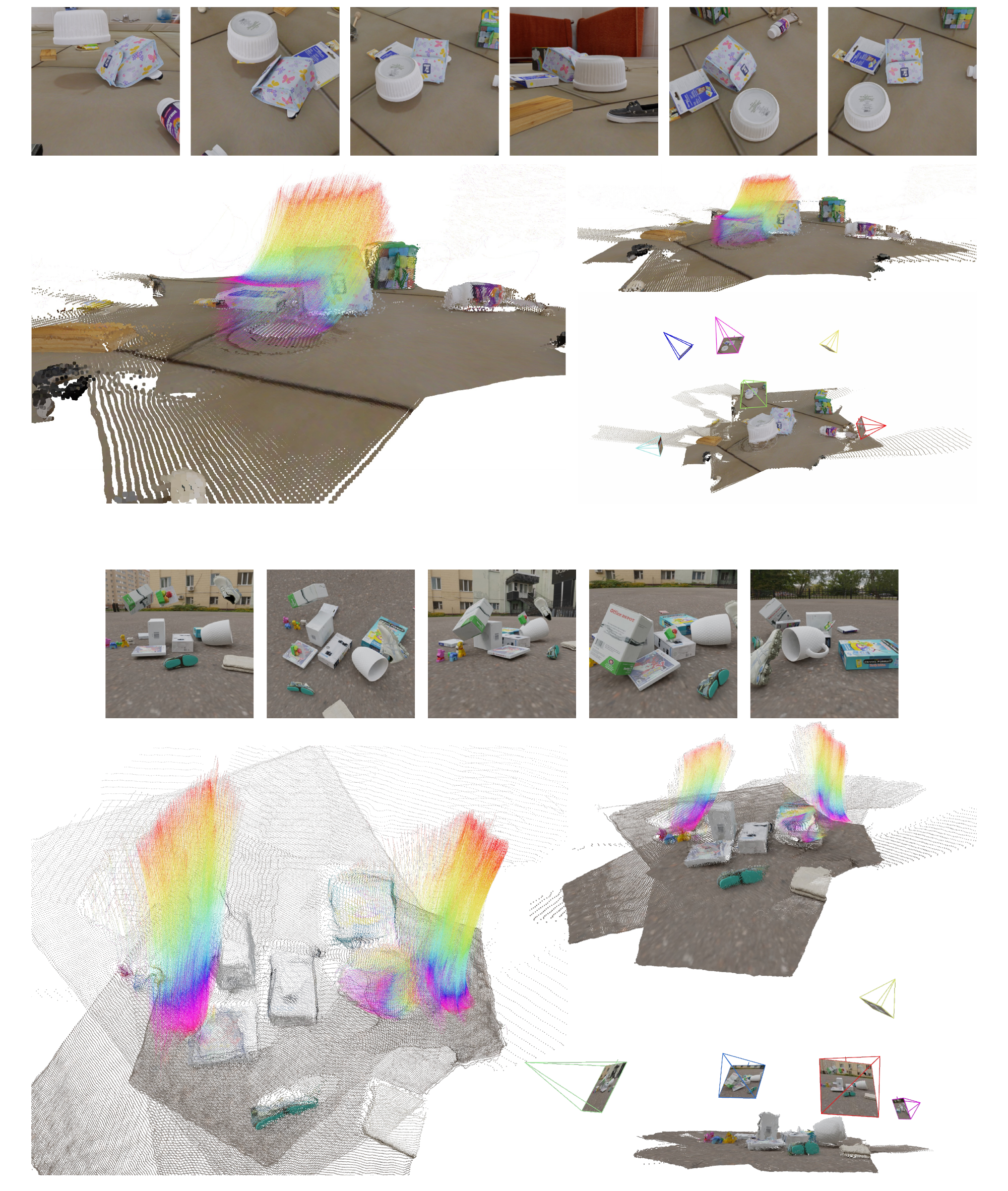}
    \caption{Trajectory fields and camera poses estimated from an unstructured, unordered image set. No sequence information is provided to the model.}
    \label{fig:unordered}
\end{figure}

\noindent\textbf{Unstructured image set to trajectory field.}  
Beyond videos or image pairs, our method also handles unstructured, unordered image sets, a setting not addressed by prior work. The inputs lack both temporal ordering and continuous camera motion, yet our framework by design can also cope with such challenging cases. As shown in \Cref{fig:unordered}, our method predicts plausible trajectory fields and camera poses under these conditions. For clarity, we present the input images in chronological order, although no sequence information is provided to the model.

\subsection{Quantitative Evaluation}
\label{sec:quantitative}

We quantitatively evaluate trajectory field estimation on the \textit{Trace Anything benchmark}, introduced in \Cref{sec:benchmark}. In contrast to established point tracking benchmarks, which evaluate trajectories only from the first frame, our protocol requires \textit{all-to-all} predictions: every pixel in every frame must be associated with a complete trajectory spanning the entire sequence. Evaluation is conducted in two settings: (i) \textit{video-based inference}, where models process 30-frame video clips, and (ii) \textit{image-pair-based inference}, where trajectories are estimated from two frames sampled 5 frames apart. All evaluations were conducted using a single NVIDIA A100 GPU. We present other quantitative results in \Cref{supp:quantitative} and ablation study in \Cref{supp:ablation}.
% We present qualitative comparisons in \Cref{sec:davis_comparison}, and additional quantitative results on an out-of-distribution dataset (\Cref{supp:pointodyssey}) as well as established 3D tracking benchmarks (\Cref{supp:3dtrack}).

\noindent\textbf{Metrics.}  
We evaluate reconstruction accuracy using end-point error (EPE). Specifically, EPE$_\text{mix}$, EPE$_\text{sta}$, and EPE$_\text{dyn}$ measure the average 3D end-point error over all points, static points, and dynamic points, respectively. To further verify whether the conditions C1 and C2 introduced in \Cref{sec:methods} are satisfied, we introduce two complementary metrics. \textit{Static Degeneracy Deviation (SDD)} quantifies the temporal jitter of trajectories in static regions, where smaller values indicate better compliance with C1. \textit{Correspondence Agreement (CA)} measures how consistently dynamic trajectories are predicted from corresponding pixels in different source frames, with lower values indicating better compliance with C2.  

\noindent\textbf{Baselines.}  
For video-based inference, we compare against state-of-the-art dynamic scene reconstruction and point tracking approaches, including CoTracker3~\citep{karaev2024cotracker3} (lifted to 3D using VGGT~\cite{wang2025vggt}), DELTA~\citep{ngo2024delta}, SpaTrackerV2~\citep{xiao2025spatialtrackerv2}, 
MonsT3R~\citep{zhang2024monst3r},
St4RTrack~\citep{feng2025st4rtrack}, POMATO~\citep{zhang2025pomato} and Easi3R~\citep{chen2025easi3r}. For image-pair inference, we compare against the optical flow method SEA-RAFT~\citep{wang2024sea} (lifted to 3D with VGGT~\cite{wang2025vggt}), the scene flow method RAFT-3D~\citep{teed2021raft}, and several 3D reconstruction approaches, including MASt3R~\citep{leroy2024grounding}, MonST3R~\citep{zhang2024monst3r}, St4RTrack~\citep{feng2025st4rtrack}, POMATO~\citep{zhang2025pomato} and Easi3R~\citep{chen2025easi3r}.

\noindent\textbf{Results.}
Quantitative results are shown in \Cref{tab:our_benchmark_video,tab:our_benchmark_pair}. Trace Anything achieves the best performance across all metrics, substantially reducing end-point errors in both static and dynamic regions while also attaining the lowest SDD and CA, indicating stronger compliance with consistency conditions. Moreover, it runs over an order of magnitude faster than optimization-based approaches, underscoring the advantage of our one-pass design.  
\begin{table}[t]
\centering
\caption{\textbf{Quantitative results on video-based trajectory field estimation. }
CA is reported in $10^{-2}$ and SDD in $10^{-3}$. Best in \textbf{bold}, second-best \underline{underlined}.}
\label{tab:our_benchmark_video}
\resizebox{0.8\linewidth}{!}{
\setlength{\tabcolsep}{6pt}
\begin{tabular}{lcccccc}
\toprule
Method & EPE$_\text{mix}$ $\downarrow$ & EPE$_\text{sta}$ $\downarrow$ & EPE$_\text{dyn}$ $\downarrow$ & CA $\downarrow$ & SDD$\downarrow$ & Runtime (s)$\downarrow$ \\
\midrule
CoTracker3 + VGGT & 0.518 & 0.461 & 0.555 & 7.83 & 1.67 & 197.4 \\
DELTA             & 0.404 & 0.384 & 0.425 & 6.24 & 1.75 & 231.6 \\
SpaTrackerV2      & 0.296 & 0.291 & 0.366 & 7.24 & 1.51 & 178.4 \\
MonsT3R & 0.316 & 0.258 & 0.330 & 8.77 & 1.74 & 99.1 \\
St4RTrack    & 0.278 & \underline{0.247} & 0.370 & 9.37 & 1.76 & \underline{22.5} \\     
POMATO   & \underline{0.272} & 0.254 & \underline{0.308}         
 & 6.78 &  \underline{1.44}  & 81.8 \\
Easi3R     & 0.308 & 0.302 & 0.324       
 & \underline{5.15} & 1.55 & 130.9 \\
\midrule
Trace Anything    & \textbf{0.234} & \textbf{0.218} & \textbf{0.295} & \textbf{5.09} & \textbf{1.06} & \textbf{2.3} \\
\bottomrule
\end{tabular}
}
\end{table}

\begin{table}[t]
\centering
\caption{\textbf{Quantitative results on image-pair-based trajectory field estimation.} 
CA is reported in $10^{-2}$ and SDD in $10^{-3}$. Best in \textbf{bold}, second-best \underline{underlined}.}
\label{tab:our_benchmark_pair}
\resizebox{0.8\linewidth}{!}{
\setlength{\tabcolsep}{6pt}
\begin{tabular}{lcccccc}
\toprule
Method & EPE$_\text{mix}$ $\downarrow$ & EPE$_\text{sta}$ $\downarrow$ & EPE$_\text{dyn}$ $\downarrow$ & CA $\downarrow$ & SDD$\downarrow$ & Runtime (s)$\downarrow$ \\
\midrule
SEA-RAFT + VGGT   & 0.226 & 0.193 & 0.427 & 18.22 & 0.77 & 1.91 \\
RAFT-3D           & 0.281 & 0.219 & 0.324 & 17.50 & 0.98 & \underline{0.37} \\
MASt3R            & 0.220 & 0.181 & {0.328} & 33.99 & 1.70 & 2.39 \\
MonST3R           & 0.206 & 0.167 & 0.346 & 20.10 & 1.25 & 2.51 \\
St4RTrack         & 0.211 & 0.202 & 0.325 & \underline{15.33} & \underline{0.63} & 1.39 \\
POMATO            & \underline{0.181} & \underline{0.137} & \underline{0.320} & 19.58 & 0.84 & 4.75 \\
Easi3R            & 0.284 & 0.269 & 0.323 & 20.41 & 0.91 & 5.08 \\
\midrule
Trace Anything    & \textbf{0.135} & \textbf{0.106} & \textbf{0.304} & \textbf{12.41} & \textbf{0.54} & \textbf{0.20} \\
\bottomrule
\end{tabular}
}
\end{table}

\begin{figure}[t]

  \centering
  \begin{minipage}{0.34\linewidth}
    \centering
    \includegraphics[width=\linewidth]{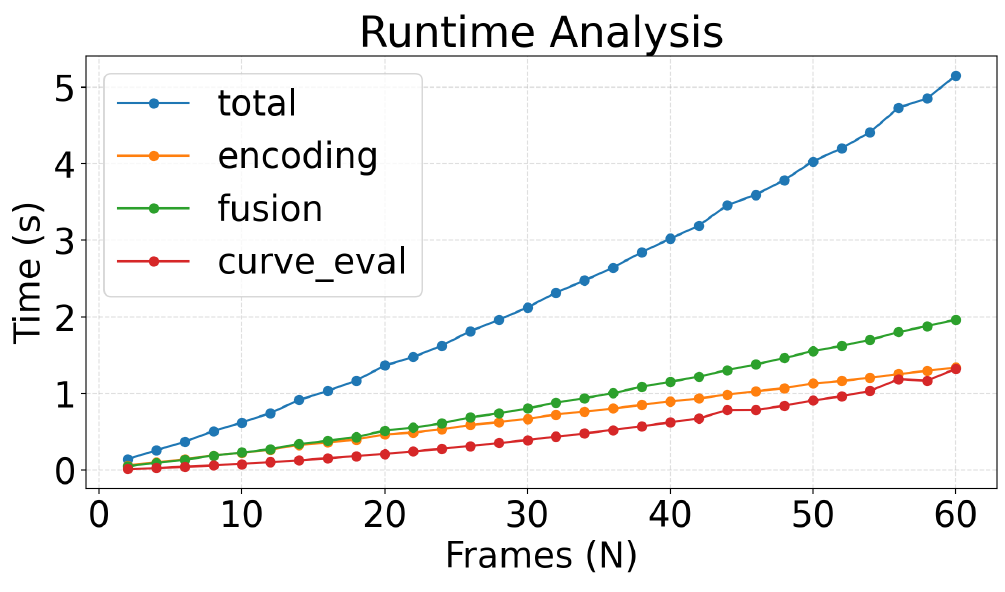}
    \caption{Stage-wise runtime vs. number of input frames.}
    \label{fig:timings}
  \end{minipage}\hfill
  \begin{minipage}{0.64\linewidth}
    \centering
    \includegraphics[width=\linewidth]{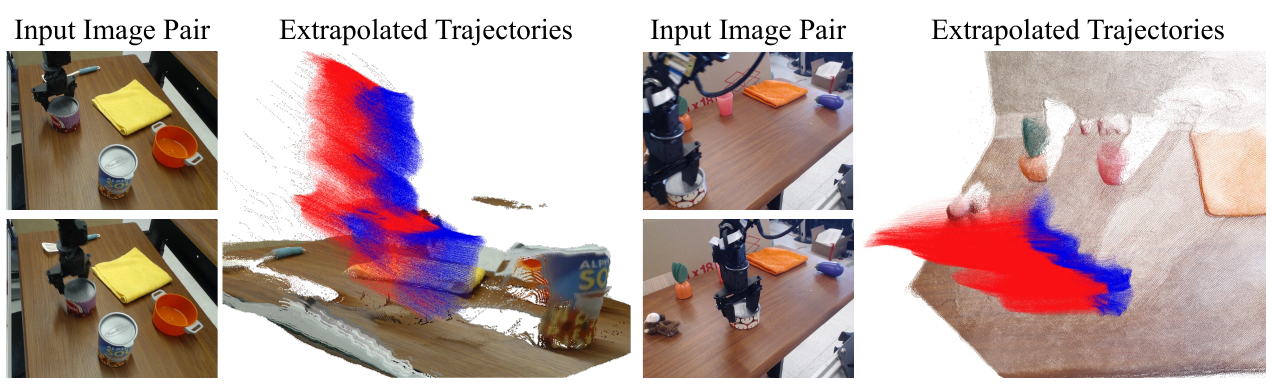}
    \caption{\textbf{Velocity-based forecasting.} 
      Per-pixel trajectories are extrapolated by tangent continuation, 
      with reconstructed trajectories in red and extrapolated ones in blue.}
    \label{fig:extra}
  \end{minipage}
\end{figure}

\noindent\textbf{Runtime breakdown.} As shown in \Cref{fig:timings}, our approach exhibits a total runtime that scales approximately linearly with the number of frames. The fusion transformer is the most time-consuming stage, followed by image encoding and curve evaluation. With single-pass inference and no per-scene optimization or external estimators, it exhibits a clear efficiency advantage, as shown in \Cref{tab:our_benchmark_video,tab:our_benchmark_pair}.

\begin{figure*}[t]
    \centering
    \includegraphics[width=\linewidth]{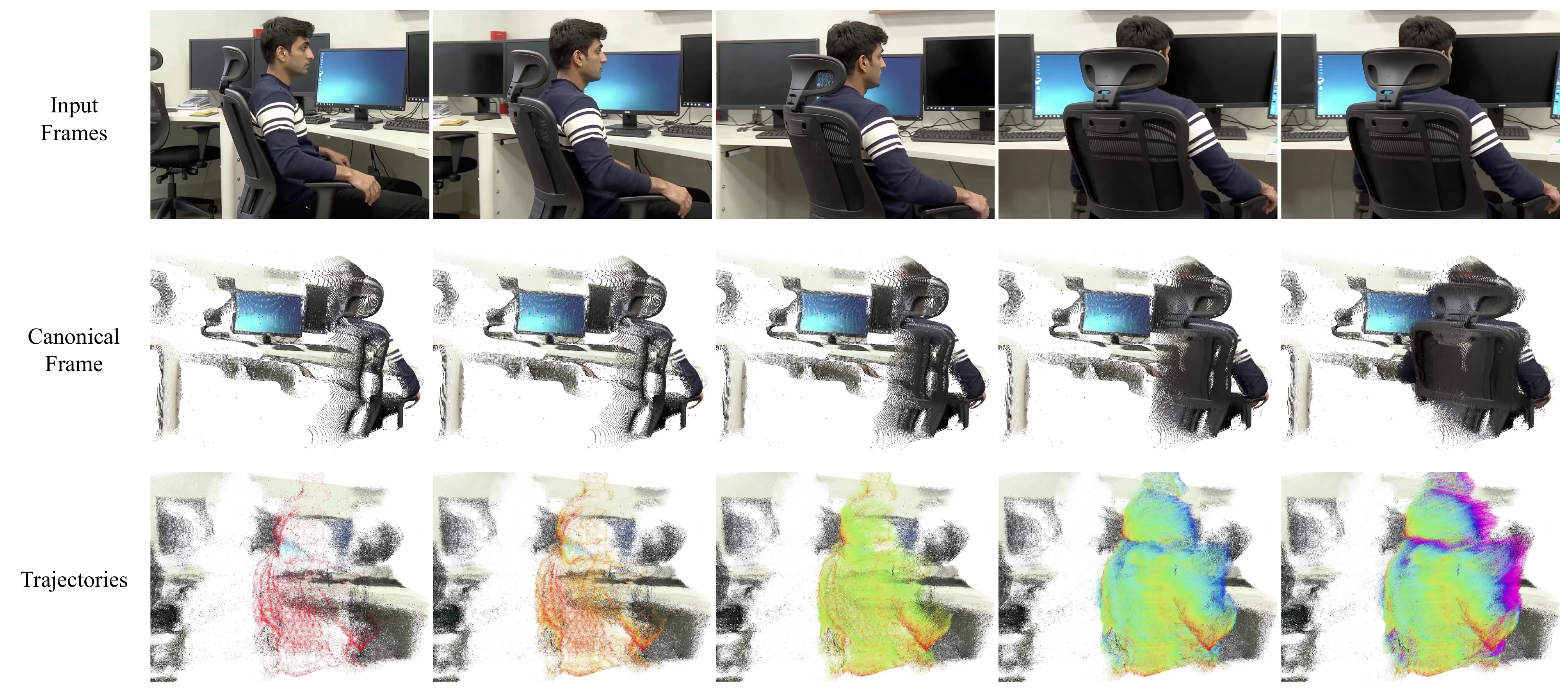}\\
    \caption{\textbf{Spatio-temporal fusion.} The trajectory field can be leveraged to fuse observations of the dynamic entity across different frames into a canonical frame.}
    \label{fig:fusion}
\end{figure*}

\subsection{Emergent Capabilities}
\label{sec:emergent}

Trajectory Field representation and Trace Anything model exhibit emergent capabilities that competing approaches do not support.

\noindent\textbf{Velocity-based forecasting.}  
The trajectory field inherently encodes 3D point velocities, per-pixel trajectories can be extrapolated by tangent continuation, allowing dense motion forecasting without additional predictors, as shown in \Cref{fig:extra}.

\begin{figure}[H]
    \centering
    \includegraphics[width=\linewidth]{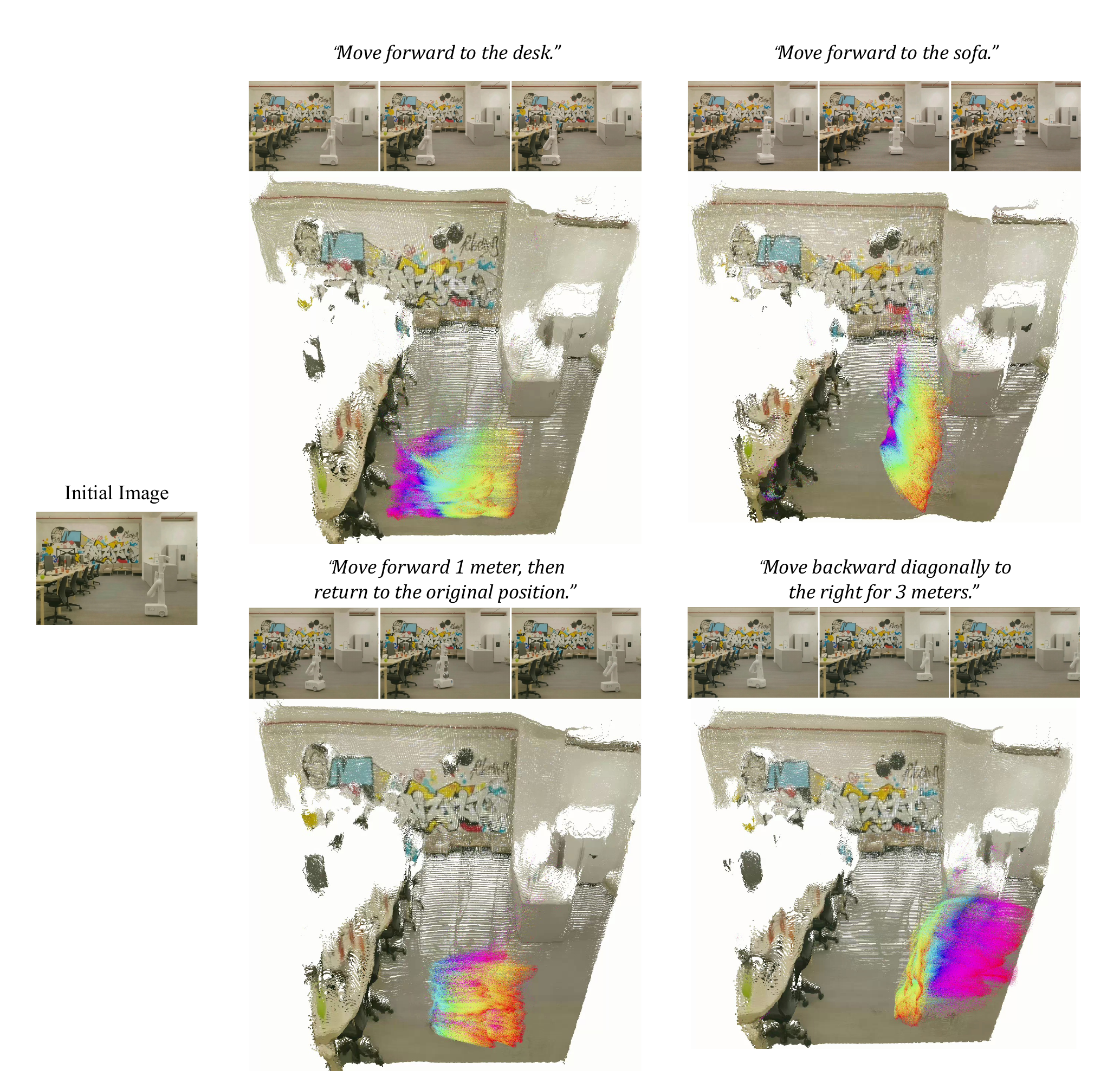}
    \caption{Instruction-based forecasting.
    Future states are generated using Seedance 1.0.}
    \label{fig:robot_choice}
\end{figure}

\noindent\textbf{Instruction-based forecasting.} With natural language instructions as input, we leverage image or video generation models to produce future states conditioned on the instructions, and then apply Trace Anything to lift these forecasts into 2D trajectory fields.  
% In \Cref{fig:fruit}, we forecast hand–object interactions under different instructions. We use Gemini 2.5 Flash Image (Nano Banana) to generate images of future states corresponding to each instruction, and then apply Trace Anything on the generated image pairs to estimate the trajectory field.  
In \Cref{fig:robot_choice}, we forecast robot actions conditioned on different instructions. We use Seedance 1.0~\citep{gao2025seedance} to generate videos of future states for different instructions, and then apply Trace Anything to predict the trajectory fields from the generated videos.  

% \begin{figure}[t]
%     \centering
%     \includegraphics[width=\linewidth]{figure/fruit.pdf}
%     \caption{Instruction-based forecasting.
%     Future states are generated using Nano Banana.}
%     \label{fig:fruit}
% \end{figure}

\noindent\textbf{Spatio-temporal fusion.}
In \Cref{fig:fusion}, predicted trajectory fields enable dynamic entities observed across multiple frames to be consistently fused back into a common canonical frame. This provides a mechanism for aggregating partial observations over time, overcoming occlusions and view changes by aligning pixels to a common reference.

\section{Conclusion}

\label{sec:conclusion}
We introduced \textbf{Trajectory Fields}, a 4D representation that encodes each pixel of each frame as a 3D trajectory, and \textbf{Trace Anything}, a feed-forward model that predicts trajectory fields from input frames, eliminating auxiliary estimators and per-scene optimization. To support large-scale learning and evaluation, we developed a synthetic data platform. Experiments show that Trace Anything delivers competitive accuracy and inference efficiency, while exhibiting new capabilities.

{
\bibliographystyle{plainnat}
\def\CVPR{IEEE/CVF Conference on Computer Vision and Pattern Recognition (CVPR)}\def\ECCV{ European Conference on Computer Vision (ECCV)}\def\ICCV{IEEE/CVF International Conference on Computer Vision (ICCV)}\def\NIPS{Advances in Neural Information Processing Systems (NeurIPS)}\def\ICML{International Conference on Machine Learning (ICML)}\def\ICLR{International Conference on Learning Representations (ICLR)}\def\WACV{IEEE/CVF Winter Conference on Applications of Computer Vision (WACV)}\def\CVPRW{IEEE/CVF Conference on Computer Vision and Pattern Recognition (CVPR) Workshops}\def\ICCVW{IEEE/CVF International Conference on Computer Vision (ICCV) Workshops}\def\ICRA{IEEE International Conference on Robotics and Automation (ICRA)}\def\TOG{ACM Transactions on Graphics (TOG)}\def\PAMI{IEEE Transactions on Pattern Analysis and Machine Intelligence (PAMI)}\def\TIP{IEEE Transactions on Image Processing (TIP)}\def\IJCV{International Journal of Computer Vision (IJCV)}\def\SIGGRAPH{ACM Transactions on Graphics
  (SIGGRAPH)}\def\SIGGRAPHASIA{ACM Transactions on Graphics (SIGGRAPH Asia)}\def\TOG{ACM Transactions on Graphics (TOG)}\def\threedv{International Conference on 3D Vision (3DV)}\def\TVCG{IEEE Transactions on Visualization and Computer Graphics (TVCG)}\def\PMLR{Proceedings of Machine Learning Research (PMLR)}

}

\clearpage

\beginappendix

\renewcommand\thefigure{\Alph{figure}} % redefining the figure numbering style
\renewcommand\thetable{\Alph{table}}   % redefining the table numbering style
\renewcommand\thesection{\Alph{section}}   % redefining the table numbering style
\setcounter{figure}{0} % reset counter 
\setcounter{table}{0} % reset counter
\setcounter{section}{0} % reset counter

% Defining the appendix section for B-spline implementation details

\tableofcontents

\section{Fields}
\label{supp:field}
A \textit{field} is a mapping defined on a domain \(M\), either a continuous space (e.g., \(\mathbb{R}^n\)) or a discrete space (e.g., \(\mathbb{Z}^n\)), to a codomain \(V\), which may be a scalar space (e.g., \(\mathbb{R}\)), a vector space (e.g., \(\mathbb{R}^3\)), or a function space (e.g., \(C^\infty(N)\)). Formally, the field is given by
\begin{equation}
    \mathcal{F}: M \to V.
\end{equation}
For instance, the radiance field introduced in \citep{mildenhall2020nerf} maps a 3D coordinate \(\mathbf{x} \in \mathbb{R}^3\) and a viewing direction \(\mathbf{d} \in S^2\) (the unit sphere) to a density \(\sigma \in \mathbb{R}^+\) and a color \(\mathbf{c} \in \mathbb{R}^3\). This is expressed as
\begin{equation}
\begin{aligned}
    \mathcal{R}: \mathbb{R}^3 \times S^2 &\to \mathbb{R}^+ \times \mathbb{R}^3, \\
    (\mathbf{x}, \mathbf{d}) &\mapsto (\sigma, \mathbf{c}).
\end{aligned}
\end{equation}
As discussed in \Cref{sec:problem}, the trajectory field introduced in this work is defined as
\begin{equation}
\begin{aligned}
   \mathcal{T}: [N] \times [H] \times [W] &\to C([0,1], \mathbb{R}^3), \\
   (i, u, v) &\mapsto \mathbf{x}_{i,u,v}(\cdot),
\end{aligned}
\end{equation}
where \([N]\), \([H]\), and \([W]\) denote the discrete sets of frame indices and pixel coordinates, respectively, and \(\mathbf{x}_{i,u,v} : [0,1] \to \mathbb{R}^3\) is a continuous 3D trajectory for pixel \((u,v)\) in frame \( I_i \). 
The domain is \( M = [N] \times [H] \times [W] \), and the codomain is \( V = C([0,1], \mathbb{R}^3) \), the space of continuous functions from \([0,1]\) to \(\mathbb{R}^3\).

\section{Parametric Curves}
\label{supp:parametric_curves}
Our trajectory field representation assigns each pixel to a 3D trajectory, expressed as a parametric curve.
In computer graphics, parametric curves are essential for modeling smooth trajectories and surfaces in applications like geometric design and animation~\citep{farin2002curves,piegl1997nurbs}. A spline-based parametric curve \(\mathbf{x}(t) : [0,1] \to \mathbb{R}^3\) maps a parameter \( t \in [0,1] \) to 3D space, defined as
\begin{equation}
    \mathbf{x}(t) = \sum_{k=0}^{n-1} \mathbf{P}_k \phi_k(t),
\end{equation}
where \(\mathbf{P}_k \in \mathbb{R}^3\) are control points and \(\phi_k(t)\) are basis functions.

As a widely used class, B\'ezier curves use Bernstein polynomials as basis functions. A B\'ezier curve of degree \( d \) with \( d+1 \) control points \(\mathbf{P}_0, \mathbf{P}_1, \dots, \mathbf{P}_d \in \mathbb{R}^3\) is defined as
\begin{equation}
    \mathbf{x}(t) = \sum_{i=0}^d \mathbf{P}_i B_{i,d}(t), \quad B_{i,d}(t) = \binom{d}{i} t^i (1-t)^{d-i},
\end{equation}
where \( B_{i,d}(t) \) are Bernstein polynomials~\citep{farin2002curves}. B\'ezier curves interpolate the first and last control points but lack local control, as adjusting one control point affects the entire curve.

B-spline curves, in contrast, provide local control through a knot vector that defines the parameter intervals where basis functions are active. A B-spline curve of degree \( p \) with control points \(\mathbf{P}_0, \mathbf{P}_1, \dots, \mathbf{P}_{n-1} \in \mathbb{R}^3\) is defined as
\begin{equation}
    \mathbf{x}(t) = \sum_{i=0}^{n-1} \mathbf{P}_i N_{i,p}(t),
\end{equation}
where \( N_{i,p}(t) \) are B-spline basis functions determined by a knot vector via the Cox-de Boor recursion formula~\citep{deboor1978practical}. 

% For a specific B-spline formulation used in our implementation, see Appendix~\ref{sec:supp_bspline}.
% \section{B-Spline Implementation for Trajectory Fields}
% \label{sec:supp_bspline}

In our implementation of Trace Anything, we employ cubic B-splines (\( p=3 \)) with clamped, non-uniform knot vectors to parameterize trajectories \(\mathbf{x}_{i,u,v}(t)\). Each segment is defined by four control points, corresponding to the cubic degree (\( p=3 \)). A trajectory is defined as
\begin{equation}
    \mathbf{x}_{i,u,v}(t) = \sum_{k=0}^{n-1} \mathbf{P}^{(k)}_{i,u,v} N_{k,3}(t),
\end{equation}
where \(\mathbf{P}^{(k)}_{i,u,v} \in \mathbb{R}^3\) are control points indexed by \(i, u, v\), and \(N_{k,3}(t)\) are cubic B-spline basis functions determined by a knot vector \(\mathbf{t} = [t_0, t_1, \dots, t_{m-1}]\). The basis functions are computed via the Cox-de Boor recursion formula:
\begin{equation}
    N_{k,0}(t) = 
    \begin{cases} 
        1 & \text{if } t_k \leq t < t_{k+1} \text{ for } k < n + p, \\
        1 & \text{if } t_k \leq t \leq t_{k+1} \text{ for } k = n + p, \\
        0 & \text{otherwise},
    \end{cases}
\end{equation}
\begin{equation}
    N_{k,p}(t) = \frac{t - t_k}{t_{k+p} - t_k} N_{k,p-1}(t) + \frac{t_{k+p+1} - t}{t_{k+p+1} - t_{k+1}} N_{k+1,p-1}(t),
\end{equation}
for \( p = 1, 2, 3 \), with non-zero denominators assumed. For \( n=4, 7, 10 \) control points, we define knot vectors with multiplicity 4 at \( t=0 \) and \( t=1 \) to ensure interpolation of the first and last control points (\(\mathbf{x}_{i,u,v}(0) = \mathbf{P}^{(0)}_{i,u,v}\), \(\mathbf{x}_{i,u,v}(1) = \mathbf{P}^{(n-1)}_{i,u,v}\)). The knot vectors \(\mathbf{t}_n\) are defined as:
\begin{equation}
    \mathbf{t}_n =
    \begin{cases}
        [0, 0, 0, 0, 1, 1, 1, 1] & \text{if } n = 4, \\
        [0, 0, 0, 0, 0.5, 0.5, 0.5, 1, 1, 1, 1] & \text{if } n = 7, \\
        [0, 0, 0, 0, 1/3, 1/3, 1/3, 2/3, 2/3, 2/3, 1, 1, 1, 1] & \text{if } n = 10.
    \end{cases}
\end{equation}
Internal knots have multiplicity up to 3, ensuring \( C^0 \)-continuity between segments. Knot differences are precomputed for efficient evaluation. Confidence values are interpolated alongside 3D coordinates \(\mathbf{P}^{(k)}_{i,u,v}\) using the same basis functions \(\phi_k(t) = N_{k,3}(t)\), enabling uncertainty-aware trajectory modeling.

\section{Additional Experimental Results}

In this section, we present additional experimental results. Please also refer to the supplementary materials for video results, including the presented features, interactive visualization demos, and qualitative comparisons.

\subsection{2D Trajectories, Dynamic Masks, Scene Flow, and Camera Poses}
\label{supp:2dtraj_etc}
The outputs of Trace Anything can naturally yield 2D trajectories, dynamic masks, scene flow, and camera poses.

\begin{figure}[h]
    \centering
    \includegraphics[width=\linewidth]{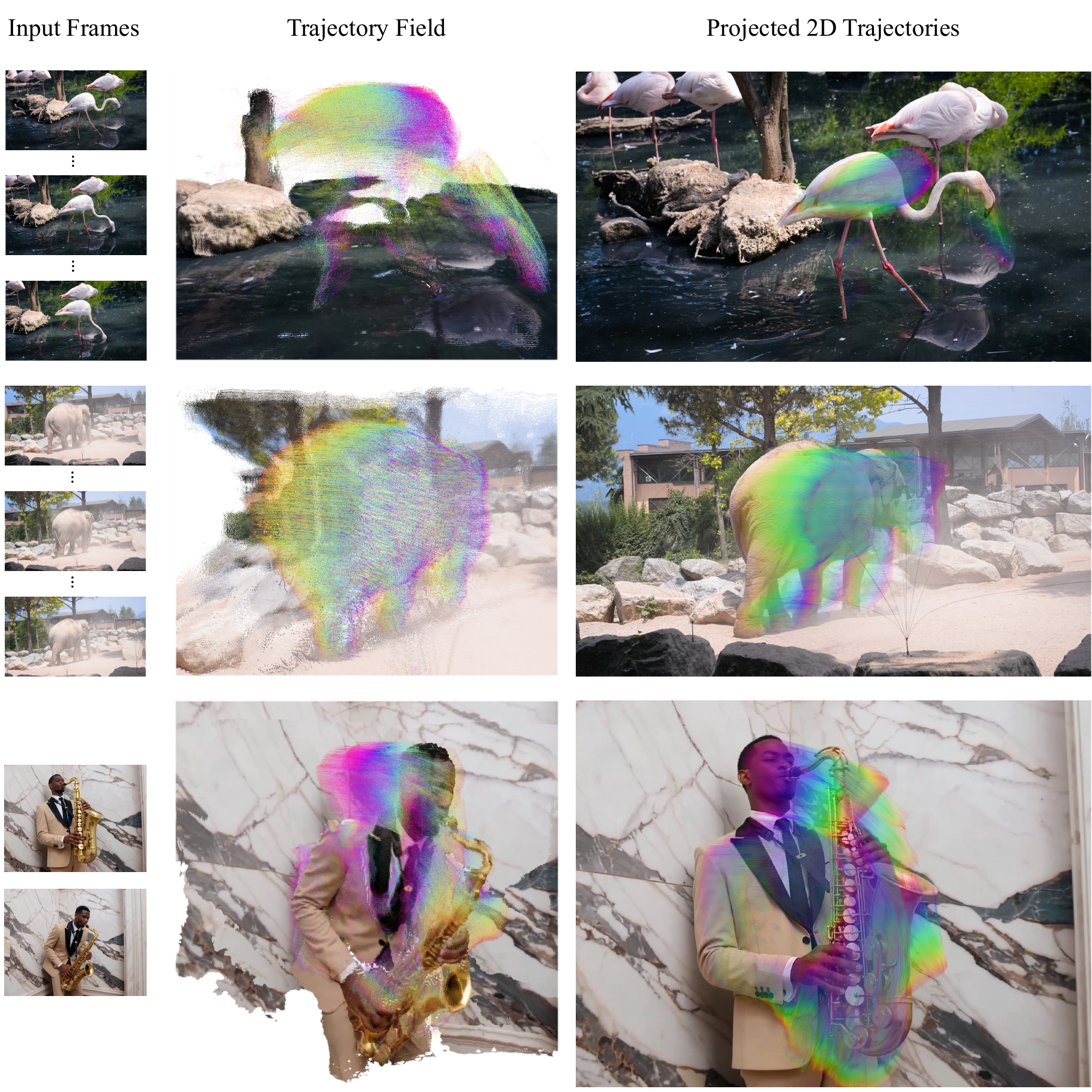}
    \caption{Projected 2D trajectories overlaid on the first input frame.}
    \label{fig:2dtraj}
\end{figure}

\noindent\textbf{2D trajectories.} Given the predicted per-pixel 3D trajectories, and with known or estimated camera parameters, we can project them into the image plane to obtain 2D trajectories. In \Cref{fig:2dtraj}, we overlay the projected 2D trajectories on the first input frame.  We also demonstrate this feature in \Cref{fig:goal}.

\begin{figure}[h]
    \centering
    \includegraphics[width=\linewidth]{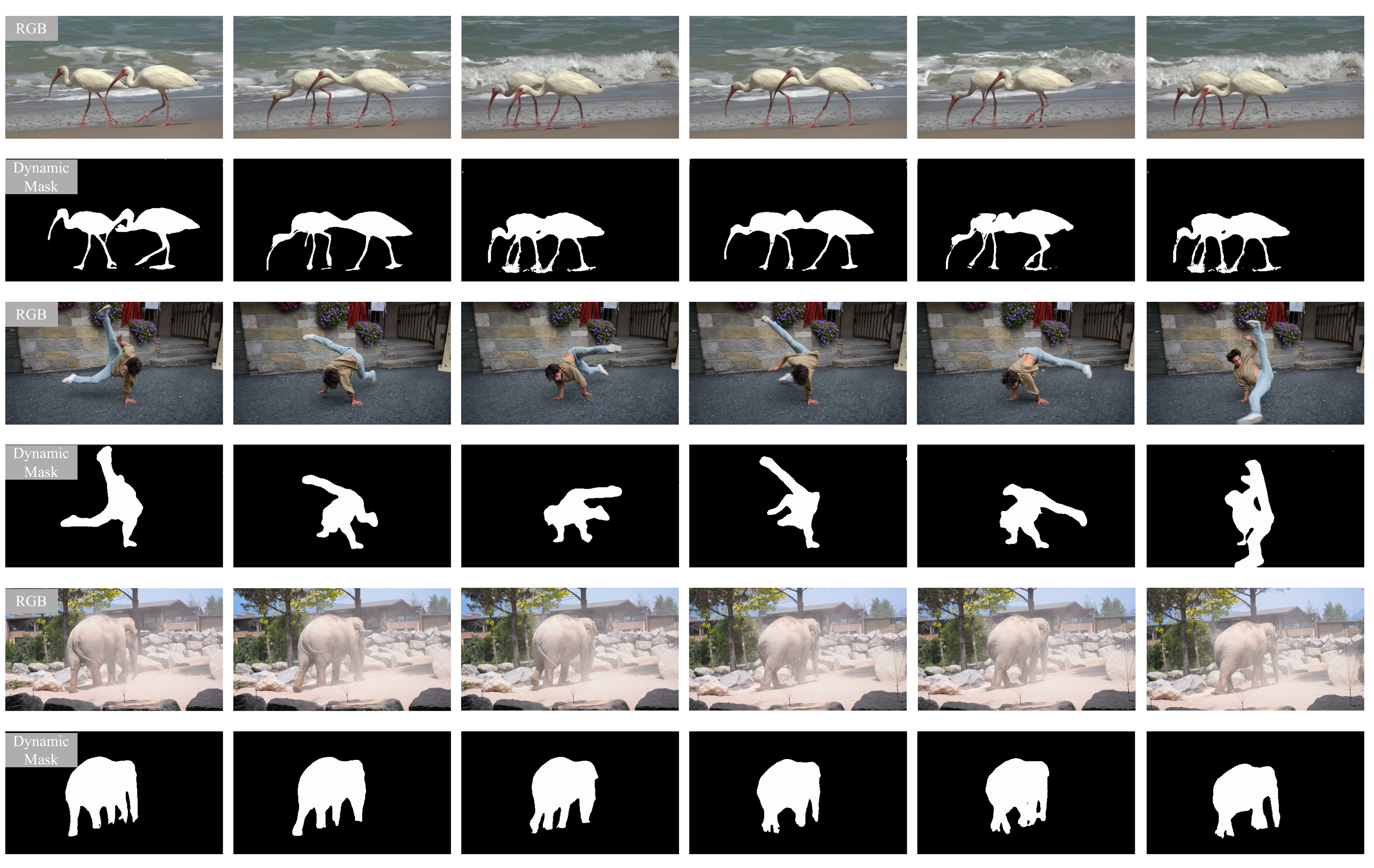}
    \caption{Dynamic mask estimation.}
    \label{fig:mask}
\end{figure}

\noindent\textbf{Dynamic masks.}
Our method effectively disentangles static and dynamic components. After Trace Anything predicts control points, we compute the variance over the control-point set associated with each pixel; thresholding this per-pixel variance yields a dynamic mask that cleanly separates static from dynamic regions, as illustrated in \Cref{fig:mask}.

\begin{figure}[h]
    \centering
    \includegraphics[width=\linewidth]{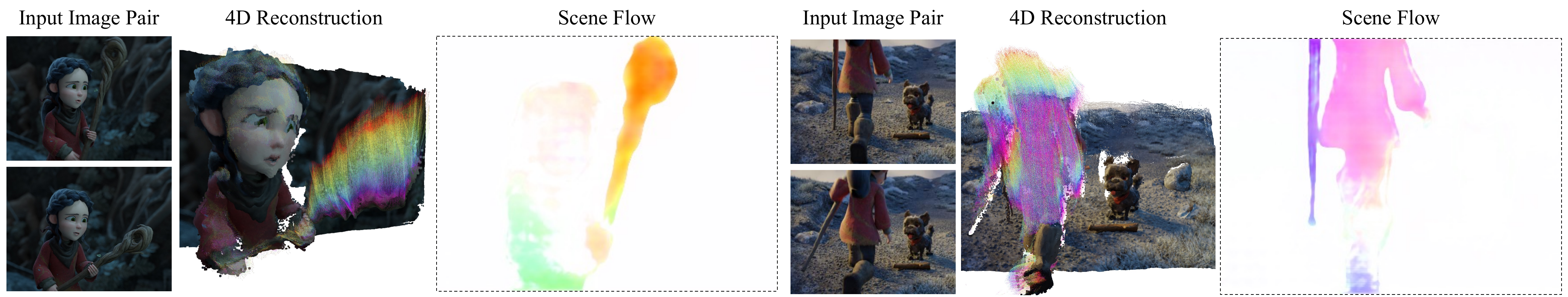}
    \caption{\textbf{4D reconstruction and scene flow from a single image pair.} From a non-consecutive image pair in the \textit{Spring} dataset, our method recovers both the 4D reconstruction and the scene flow, with $x$ and $z$ components color-coded for visualization.}
    \label{fig:spring}
\end{figure}

\noindent\textbf{Scene flow.} Given an input image pair, the scene flow can be obtained as the difference between the two endpoints of the predicted trajectories. In \Cref{fig:spring}, we present a 4D reconstruction together with the estimated scene flow from an image pair in the \textit{Spring} dataset. To highlight robustness under long-range motion, the two images are chosen from non-consecutive frames.

\begin{figure}[h]
    \centering
    \includegraphics[width=\linewidth]{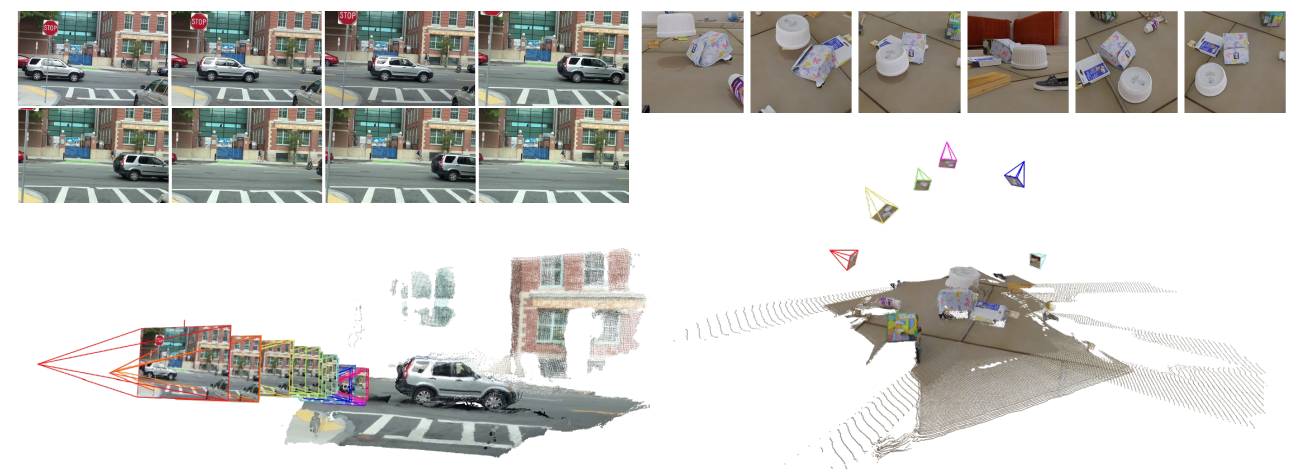}
   \caption{Estimated camera poses over the 4D reconstruction.}
    \label{fig:campose}
\end{figure}

\noindent\textbf{Camera poses.} Since \Cref{eqn:self_pointmap} provides a world-coordinate point map for each image, we follow \citet{fast3r} (Sec.~4.2) to estimate focal length, rotation, and translation. Our method handles both continuous camera motion in videos and discrete poses from unordered image sets. As shown in \Cref{fig:campose}, it correctly recovers camera motion even in dynamic scenes—for example, forward camera movement with perpendicular object motion, or objects in free fall captured by an unordered image set. In the second example, we present the input images in chronological order, although no temporal information is provided to the model.

\subsection{Qualitative Comparison}
\label{sec:davis_comparison}
\begin{figure}[ht]
    \centering
    \includegraphics[width=\linewidth]{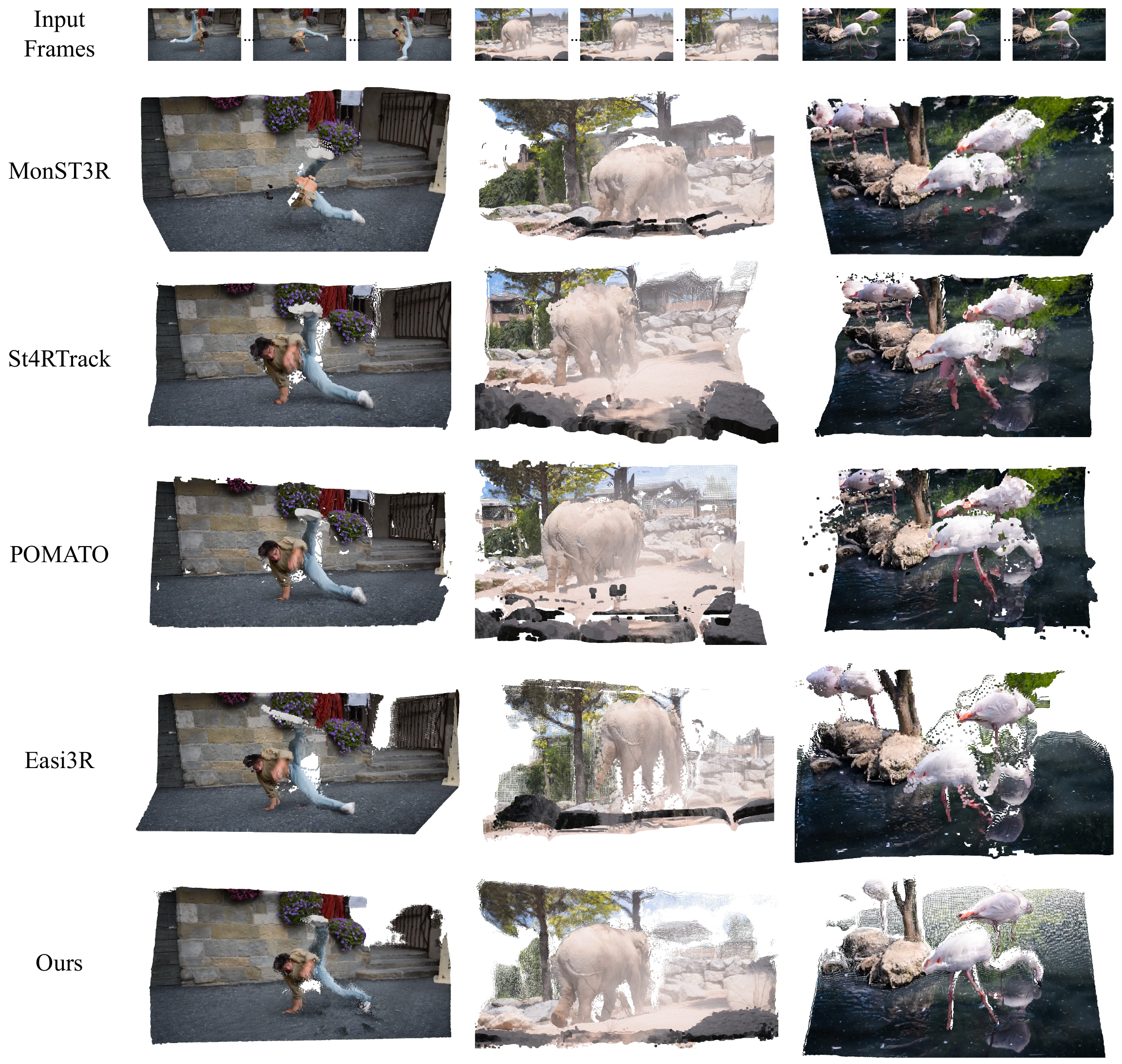}
    \caption{\textbf{Qualitative comparison on DAVIS~\citep{perazzi2016benchmark}.} 
    Our method better recovers fine details and handles complex motion while disentangling static and dynamic objects.}
    \label{fig:davis_comparison}
\end{figure}

We provide qualitative comparisons of reconstructed point clouds on DAVIS~\citep{perazzi2016benchmark}. As shown in \Cref{fig:davis_comparison}, our method better preserves fine object details (e.g., the elephant’s tail and the flamingo’s neck), correctly handles complex motion, and disentangles static and dynamic objects. Please refer to the supplementary videos for clearer visual comparisons.

\subsection{Additional Quantitative Comparison}
\label{supp:quantitative}
\noindent\textbf{Out-of-distribution input.}
To evaluate our model under out-of-distribution conditions, we construct an additional benchmark from PointOdyssey~\citep{zheng2023pointodyssey}, consisting of 50 videos of 30 frames each. Our model has never been trained or fine-tuned on PointOdyssey. As shown in \Cref{tab:pointodyssey}, our method maintains advantages across all metrics as well as inference efficiency.

\begin{table}[h]
\centering
\caption{\textbf{Quantitative results on out-of-distribution data.}
CA is reported in $10^{-2}$ and SDD in $10^{-3}$. Best in \textbf{bold}, second-best \underline{underlined}.}
\label{tab:pointodyssey}
\resizebox{0.8\linewidth}{!}{
\setlength{\tabcolsep}{6pt}
\begin{tabular}{lcccccc}
\toprule
Method & EPE$_\text{mix}$ $\downarrow$ & EPE$_\text{sta}$ $\downarrow$ & EPE$_\text{dyn}$ $\downarrow$ & CA $\downarrow$ & SDD$\downarrow$ & Runtime (s)$\downarrow$ \\
\midrule
St4RTrack    & \underline{0.269} & \underline{0.243} & \underline{0.325} & 9.82 & \underline{1.70} & \underline{19.9} \\     
POMATO   & 0.344 & 0.319 & 0.397        
 & \underline{6.24 }&  1.72  & 84.1 \\
Easi3R     & 0.368 & 0.307 & 0.376     
 & 7.10 & 1.99 & 125.1 \\
\midrule
Trace Anything    & \textbf{0.256} & \textbf{0.212} & \textbf{0.319} & \textbf{6.19} & \textbf{1.37} & \textbf{2.3} \\
\bottomrule
\end{tabular}
}
\end{table}

\noindent\textbf{3D tracking.}
Although our primary task is trajectory field estimation, our method achieves strong results on 3D tracking without task- or dataset-specific fine-tuning. We quantitatively compare against other approaches on the TAPVid-3D~\citep{koppula2024tapvid} benchmark. For each subset of TAPVid-3D (ADT, DriveTrack, and PStudio), we sample 50 videos of 60 frames each, using every other frame as input, and report APD$_{3D}$ (average percent of points within a threshold, measuring spatial accuracy) and AJ (average Jaccard, capturing both spatial and occlusion correctness).

\begin{table}[h]
\centering
% spatracker v2
\caption{\textbf{Quantitative results on 3D tracking. }
 Best in \textbf{bold}, second-best \underline{underlined}.}
\label{tab:tracking}
\resizebox{0.8\linewidth}{!}{
\setlength{\tabcolsep}{6pt}
\begin{tabular}{l cc cc cc c}
\toprule
& \multicolumn{2}{c}{ADT} & \multicolumn{2}{c}{DriveTrack} & \multicolumn{2}{c}{PStudio} &  \\
\cmidrule(lr){2-3}\cmidrule(lr){4-5}\cmidrule(lr){6-7}
Method & APD$_{3D}\,\uparrow$ & AJ\,$\uparrow$ & APD$_{3D}\,\uparrow$ & AJ\,$\uparrow$ & APD$_{3D}\,\uparrow$ & AJ\,$\uparrow$ & Runtime (s) $\downarrow$ \\
\midrule
VGGT + CoTracker & 8.9 & 9.7 & 6.2 & 5.4 & 8.6 & 5.8 & 172.4 \\
St4RTrack        & 15.2 & 13.4 & 8.5 & 7.4 & 7.2 & 6.9 &  \underline{18.9} \\
POMATO           & 18.2 & 13.6 & 11.3 & 7.8 & 12.2 & 8.3 &  69.2 \\
SpaTracker       & \underline{18.3} & \textbf{17.4} & \textbf{16.0} & \textbf{10.1} & \underline{16.2} & \underline{10.3} & 191.1 \\
\midrule
Trace Anything   & \textbf{20.5} & \underline{15.6} & \underline{15.5} & \underline{9.6} & \textbf{16.3} & \textbf{10.8} &  \textbf{2.1} \\
\bottomrule
\end{tabular}
}
\end{table}

In \Cref{tab:tracking}, we present these quantitative results. Notably, SpaTracker~\citep{xiao2024spatialtracker} is designed and trained for 3D tracking. Our approach remains competitive, surpassing it on some metrics and running orders of magnitude faster, as SpaTracker is limited to a fixed number of query points per run, whereas our model performs per-pixel tracking in a single forward pass.

\subsection{Ablation Study} 
\label{supp:ablation}
\Cref{tab:ablation} presents ablation studies on our Trace Anything benchmark, evaluating both the choice of geometric backbone and the type of parametric curve. 
For the geometric backbone, we compare the effect of initializing the image encoder and fusion transformer with different pretrained models, including Fast3R~\citep{fast3r}, VGGT~\citep{wang2025vggt}, and ``None'' (following the Fast3R architecture but with random initialization). 
For the parametric curve types, we evaluate polynomial curves\footnote{Although our method restricts parametric curves to control-point–based ones such as Bézier and B-spline, we experimented with polynomial curves during the early development phase.} as well as Bézier and B-spline curves with varying numbers of control points.

\begin{table}[h]
\centering
\caption{\textbf{Ablation study on Trace Anything benchmark.} 
CA is reported in $10^{-2}$ and SDD in $10^{-3}$.  Best in \textbf{bold}, second-best \underline{underlined}.}
\label{tab:ablation}
\resizebox{\linewidth}{!}{
\setlength{\tabcolsep}{6pt}
\begin{tabular}{l l ccccccc}
\toprule
Backbone & Curve Type & EPE$_\text{mix}$ $\downarrow$ & EPE$_\text{sta}$ $\downarrow$ & EPE$_\text{dyn}$ $\downarrow$ & CA $\downarrow$ & SDD $\downarrow$ & Runtime (s)$\downarrow$ \\
\midrule
None   & B-Spline (10 control points) & 0.472 & 0.416 & 0.505 & 8.17 & 1.08 & 2.3 \\
Fast3R & Polynomial (degree 3)      & 0.619 & 0.582 & 0.673 & 9.19 & 1.10 & 1.8 \\     
Fast3R & Bezier (4 control points)  & 0.299 & 0.271 & 0.312 & \textbf{5.08} & 1.11 & \textbf{1.7} \\
Fast3R & Bezier (10 control points) & 0.238 & 0.224 & 0.319 & 6.13 & 1.08 & 2.5\\
Fast3R & B-Spline (4 control points)  & 0.281 & 0.264 & 0.330 & 6.01& 1.08 & \textbf{1.7} \\
Fast3R & B-Spline (7 control points)  & 0.237 & 0.229 & 0.317 & 5.81 & 1.11 & 2.1 \\
Fast3R & B-Spline (10 control points) & \textbf{0.234} & \textbf{0.218} & \textbf{0.295} & \underline{5.09} & \textbf{1.06} & 2.3 \\
VGGT   & B-Spline (10 control points) & \underline{0.236} & \underline{0.221} & \underline{0.276} & 6.11 & \underline{1.07} & 7.2 \\
\bottomrule
\end{tabular}
}
\end{table}

As shown in \Cref{tab:ablation}, polynomial curves underperform because their parameters lack the clear geometric and physical interpretability. In contrast, B-spline curves with ten control points achieve the best overall performance, and accuracy generally improves as the number of control points increases.  
For the backbone, training without pretrained initialization struggles to converge. Compared with Fast3R, VGGT yields modest gains on certain metrics but incurs substantially higher runtime. Nonetheless, we observe VGGT can be beneficial in settings that demand fine structural detail or involve large-baseline scenarios. Based on these results, we adopt Fast3R with B-spline curves (10 control points) as the default configuration in Trace Anything.

\section{Limitations}
Since Trace Anything is trained for trajectory field estimation, we rely on synthetic data to obtain dense annotations. This inevitably introduces a domain gap with real-world scenarios. Incorporating partial annotations from real data may help bridge this gap and represents a promising direction for future work.  

Our parametric curve representation, with a limited number of control points, has restricted expressive power for highly complex motions. In such cases, we mitigate the issue by clipping trajectories into fixed window sizes or downsampling frames. However, these strategies may fail in scenarios such as repeated back-and-forth motion, and performance also degrades as the number of frames increases. A more fundamental solution likely requires training with larger-scale datasets with high quality.  

As the first attempt at dense per-pixel trajectory field estimation, our approach offers efficiency advantages but may be less precise than sparse 3D tracking methods~\citep{xiao2024spatialtracker, xiao2025spatialtrackerv2}. Incorporating fine-grained point-level estimation from such methods into our framework could be an interesting direction for future research.

\section{LLM Usage Declarations}
We declare that Large Language Models (LLMs) were used in a limited capacity during the preparation of this manuscript. Specifically, LLMs were employed for grammar checking, word choice refinement, and typo correction. All core technical contributions, experimental design, analysis, and conclusions are entirely our own. The use of LLMs did not influence the scientific methodology, result interpretation, or theoretical contributions of this research.

\end{document}